\documentclass[12pt]{article}
\usepackage[utf8]{inputenc}
\usepackage{lmodern}
\usepackage{amsmath,amsfonts,amssymb}
\usepackage{graphicx}
\usepackage{hyperref}
\usepackage{geometry}
\usepackage{enumitem}
\usepackage{titlesec}
\usepackage{fancyhdr}
\usepackage{setspace}
\usepackage{natbib}
\usepackage{mathrsfs}
\usepackage{authblk}
\usepackage{xcolor}
\usepackage{abstract}

\usepackage{amsthm}

\title{BEWA: A Bayesian Epistemology-Weighted Artificial Intelligence Framework for Scientific Inference}
\author{\Large Craig S. Wright\\
\small Department of Computer Science\\
\small University of Exeter\\
\small \texttt{cw881@exeter.ac.uk}}
\date{\today}

\begin{document}

\maketitle
\thispagestyle{empty}

\newpage
\begin{abstract}
\noindent The proliferation of scientific literature and the accelerating complexity of epistemic discourse have outpaced the evaluative capacities of both human scholars and conventional artificial intelligence systems. In response, we propose Bayesian Epistemology with Weighted Authority (BEWA), a computational architecture for truth-oriented knowledge modelling. BEWA formalises belief as a probabilistic relation over structured claims, indexed to authors, contexts, and replication history, and updated via evidence-driven Bayesian mechanisms. Integrating canonical authorial identification, dynamic belief networks, replication-weighted citation metrics, and epistemic decay protocols, the system constructs an evolving belief state that prioritises truth utility while resisting social and citation-based distortions. By anchoring every propositional unit in structured metadata and linking updates to semantic replication and contradiction analysis, BEWA enables automated, principled reasoning across a corpus of scientific knowledge. This work advances the theoretical foundations and practical frameworks necessary for autonomous epistemic agents to assess, revise, and propagate beliefs in dynamic scientific environments.

\end{abstract}

% Keywords (not using the keywords package)
\vspace{1em}

\newpage

\tableofcontents
\newpage

\section{Introduction}

The crisis of epistemic overload in modern scientific inquiry has exposed a critical deficiency in how truth claims are assessed, validated, and integrated across time and domain. The exponential growth in peer-reviewed publications, accompanied by inconsistent replication rates, entrenched citation biases, and the sociological entanglements of scientific authorship, has rendered traditional mechanisms of epistemic filtering increasingly obsolete. Simultaneously, artificial intelligence—while having demonstrated capacity in data correlation and language generation—remains fundamentally ill-equipped to perform rigorous epistemic reasoning. This gap is not merely technical but conceptual: current AI systems lack any principled framework for evaluating the truth-promoting value of claims, discerning authoritative sources, or understanding belief as a structured probabilistic relation between agents, claims, and contexts.

The present work introduces a formal architecture—Bayesian Epistemology with Weighted Authority (BEWA)—which systematically encodes the logic of belief formation, update, and decay, guided by the core axioms of Bayesian rationality, tempered by structural mechanisms for authority weighting, replication scoring, and temporal reassessment. BEWA does not seek to supplant human reasoning, but to embed within computational systems the formal machinery required to model, test, and evaluate propositional knowledge in scientific domains. Where contemporary models optimise for coherence or linguistic plausibility, BEWA operationalises scientific epistemology as a computable framework: every claim must be anchored, weighted, and situated within an evolving belief graph responsive to empirical reinforcement, contradiction, and decay.

At its core, this work constitutes a reimagining of what it means for AI to “know”. Rather than train a system to generate plausible continuations of language, BEWA instantiates an agent that reasons over propositions, scrutinises claims for cross-referenced coherence, and reassigns belief weightings in response to changing empirical and social contexts. In doing so, it lays the foundation for autonomous epistemic agents—not only capable of ingesting scientific knowledge, but of judging its weight, revising their stance, and constructing rational belief networks that remain resilient, transparent, and optimally truth-directed. This architecture forms not merely a tool for scientific indexing, but the cornerstone of a broader philosophical project: to make scientific reasoning computationally tractable without abandoning the rigour, fallibilism, and probabilistic humility that science demands.
\subsection{Motivation and Context}

The production of scientific literature has expanded exponentially in the last two decades, with repositories such as PubMed, arXiv, and Scopus indexing tens of millions of publications across diverse domains. While this proliferation reflects a thriving global research enterprise, it has introduced a fundamental tension: the epistemic cost of abundance. Human cognitive capacity to read, assimilate, verify, and cross-correlate claims has not scaled proportionally, leaving both researchers and knowledge systems vulnerable to the pitfalls of unreplicated findings, citation cascades, and reputational distortions. As Ioannidis (2005) and subsequent meta-analyses have shown, a significant fraction of published results—even in high-impact journals—fail to replicate, and yet are continuously cited, shaping the direction of scientific discourse in potentially misleading ways.

Conventional artificial intelligence systems, while effective at information retrieval and semantic clustering, remain epistemically shallow. Their models lack the formal apparatus to discriminate between verified knowledge, unsupported assertions, or strategic citations, and thus perpetuate the same problems of over-generalisation and under-qualification found in human discourse. The incorporation of Large Language Models (LLMs) into scientific workflows exacerbates this vulnerability, given their high fluency but epistemic indifference (Bender et al., 2021). What is needed is not a language generator, but a system for knowledge discrimination: a machine that believes rationally.

BEWA (Bayesian Epistemology with Weighted Authority) is proposed as a principled framework to address this lacuna. Drawing on foundational work in formal epistemology and probabilistic reasoning, BEWA offers a system that does not merely store scientific claims but assigns them structured belief values conditional on their provenance, corroboration, authorial record, and epistemic utility. By aligning computational epistemology with the standards of scientific methodology—replication, citation credibility, authorial accountability—BEWA is positioned to serve as a formal counterpart to scientific reasoning in silico.
\subsection{The Problem of Scientific Epistemology in AI}

The central problem of scientific epistemology in artificial intelligence is not the extraction of propositions, nor even their contextualisation, but the absence of principled belief management. Contemporary AI systems, particularly those driven by deep learning and language modelling architectures, have achieved notable success in identifying textual similarity, performing question answering, and generating plausible discursive outputs. However, these models lack any internalisation of epistemic constraints, such as the replicability of claims, the entrenchment of belief under sustained verification, or the significance of retraction and reputational decay. Without a formal epistemic scaffold, such systems merely reflect the statistical regularities of their training data, rather than critically adjudicating between levels of epistemic warrant.

The failure to incorporate epistemic stratification has deep consequences. Systems trained indiscriminately on corpora that blend rigorous research with fringe science, preprints with peer-reviewed material, or established consensus with contested speculation, conflate reliability with frequency. They lack the means to discount anomalous or outdated claims, to resolve conflicts between competing hypotheses, or to track the evolution of knowledge claims across time and replication. As such, they are incapable of supporting decision systems—scientific, legal, medical, or policy-based—that depend on epistemic stability and justified belief rather than surface-level coherence.

Moreover, epistemological naïveté in AI raises a meta-scientific hazard: the uncritical automation of flawed scientific inference. The citation of unreplicated or retracted work, the omission of contrary findings, or the aggregation of selectively reported results has long been documented as a contributor to scientific dysfunction (Ioannidis, 2005; Greenberg, 2009). When AI systems ingest and reproduce these distortions without corrective filtering, they entrench misinformation under the guise of precision. A scientific AI must not merely infer, but adjudicate—assigning belief in a manner consistent with rational constraints, probabilistic coherence, and methodological rigour. This is the foundational imperative that BEWA addresses.
\subsection{Objectives of BEWA}

The Bayesian Epistemology Weighting Architecture (BEWA) is developed to resolve the epistemic lacunae in contemporary AI systems by formalising a structured, axiomatic approach to scientific belief management. Its core objective is the construction of a truth-promoting knowledge architecture in which claims are not simply stored or retrieved, but dynamically assessed, weighted, and propagated in accordance with their epistemic warrant. To this end, BEWA integrates a full Bayesian inferential layer wherein prior beliefs are established from canonical sources, updated through reproducible evidence, and decayed with temporal distance or contradiction. The resulting framework is not static but temporally responsive, adjusting belief states in proportion to the evolving consensus and integrity of underlying claims.

BEWA's second principal objective is to embed authorial credibility and citation structure as explicit variables in belief formation. Unlike models that treat all sources equivalently, BEWA distinguishes canonical authors from peripheral contributors through reputation modelling, peer review engagement, replication track record, and retraction history. Each claim inherits a weighted score derived not only from its own properties, but from the broader epistemic ecosystem of author, venue, and reception. In this way, BEWA promotes methodological accountability: authors whose work has endured scrutiny and replication enhance the evidential force of their claims; those whose work is anomalous, inconsistent, or retracted attenuate the propagation of falsehoods.

Thirdly, BEWA is designed to operationalise these epistemic mechanisms across complex, multi-claim knowledge graphs. Through a formal propositional structure, semantic linkage, and probabilistic belief propagation, BEWA enables the dynamic evolution of belief networks wherein conflict, contradiction, and refinement are explicitly modelled. It does not seek finality, but stability under continuous critical evaluation. The architecture aims not merely to aggregate knowledge, but to support scientific reasoning—through structured epistemic decay, replication-triggered resets, and contradiction-responsive attenuation. BEWA thereby elevates AI from passive aggregator to active epistemic agent, consistent with the axioms of Bayesian rationality and the imperatives of contemporary philosophy of science.
\subsection{Contributions and Novelty}

This work introduces the Bayesian Epistemology Weighting Architecture (BEWA), a mathematically rigorous, axiomatically grounded system for autonomous epistemic processing in scientific domains. Unlike prior models that conflate data aggregation with epistemic judgment, BEWA offers a formal mechanism for belief formation, revision, and decay rooted in Bayesian inference and justified by philosophical and computational standards of rationality. Its novelty lies in the explicit coupling of structured propositional claims with a belief updating framework that is temporally sensitive, source-critical, and capable of representing both support and contradiction across a dynamically evolving knowledge space.

BEWA’s primary contribution is the integration of epistemic virtues—replicability, citation strength, author credibility, and peer review participation—into a formally specified weighting mechanism. These properties are not ad hoc features but are embedded as first-class terms within the probabilistic architecture. The system models belief not as a monolithic scalar but as a multidimensional function of authorial history, domain coherence, and cross-claim dependency. Moreover, it introduces the notion of epistemic decay and rejuvenation protocols: aged or isolated claims lose credence over time unless renewed through replication or citation; conversely, successful replications trigger weight propagation throughout associated belief clusters.

A further contribution is the development of a semantic infrastructure that enables BEWA to interpret, map, and reconcile overlapping claims across multiple disciplines and terminological systems. By constructing belief networks with semantic equivalence mappings and contradiction matrices, BEWA supports cross-disciplinary synthesis without erasing contextual nuance. Finally, the architecture’s modularity, cryptographic provenance tracking, and audit interfaces ensure reproducibility and transparency, making it suitable not only for scientific archiving but for regulatory, forensic, and policy-critical applications. In aggregating formal epistemology, machine reasoning, and information security, BEWA constitutes a foundational advancement in the design of truth-sensitive artificial intelligence.

\section{Overview of System Architecture}

This section provides a structural and conceptual overview of the Bayesian Epistemology-Weighted AI (BEWA) system. It delineates the foundational principles that govern the architectural framework, mapping the philosophical underpinnings of the system into concrete computational form. The system is not merely an exercise in engineering but a principled implementation of formal epistemology, designed to handle dynamic and conflicting scientific information within a rational, belief-updating paradigm. Accordingly, the architecture is built to ensure that knowledge claims are not merely stored or retrieved, but are actively evaluated, weighted, and revised through rigorously formalised evidentiary structures. BEWA is therefore conceived as a continuously self-correcting epistemic engine, capable of assessing propositional validity with statistical robustness, while maintaining scepticism until replicability thresholds are met.

The structure of the system rests on three critical pillars. First, the philosophical grounding in Bayesian epistemology ensures that all claims are embedded within a probabilistic inferential model, allowing beliefs to evolve incrementally and cautiously. Second, the architectural design is modular and hierarchical, integrating ingest pipelines, structured claim representation, belief propagation, and dynamic decay mechanisms, while ensuring coherence across each component. Third, the system embodies a core commitment to epistemic integrity: it prioritises replicability, resists premature belief inflation, and penalises epistemically hollow citations or uncorroborated popularity. This section thus introduces the high-level logic of BEWA’s operation, setting the stage for the detailed mechanics explored in subsequent sections.

\subsection{Philosophical Basis: Bayesian Epistemology}

\textbf{Axiom 1 (Rational Belief as Probability):} \emph{Any rational agent’s degree of belief in a proposition $H$ must be representable by a real-valued probability $P(H)$ within the closed interval $[0, 1]$, such that the agent’s belief system adheres to the Kolmogorov axioms of probability.}

This axiom forms the foundational commitment of BEWA’s inferential framework: epistemic states are mapped to probability distributions, and rational updates to these states are governed by Bayes' Theorem. This view is justified on both normative and decision-theoretic grounds. As shown in \citet{ramsey1931truth} and formalised by \citet{deFinetti1937}, the use of probabilities to represent belief is both behaviourally and logically coherent when an agent aims to avoid Dutch Book incoherence. That is, if one's belief assignments violate the probability axioms, a set of bets can be constructed that guarantees a loss, revealing the irrationality of those beliefs.

\textbf{Definition 1 (Bayesian Agent):} An agent $A$ is said to be Bayesian if for any proposition $H$ and evidence $E$, its belief update adheres to:
\[
P(H \mid E) = \frac{P(E \mid H) \cdot P(H)}{P(E)},
\]
provided $P(E) > 0$.

\textbf{Theorem 1 (Bayesian Coherence Criterion):} \emph{If an agent's beliefs are updated via Bayes’ Theorem, and their priors obey the probability axioms, then the agent is immune to Dutch Book constructions.}

\begin{proof}
See \citet{vanFraassen1989} and \citet{joyce1998}, where the coherence of Bayesian updating is proved using decision-theoretic formalism. The key idea is that any deviation from Bayes' Rule allows a clever adversary to exploit inconsistencies via bets that yield guaranteed losses.
\end{proof}

BEWA adopts the Bayesian framework not as a mere computational convenience, but as a foundational epistemological commitment. Within the philosophy of science, Bayesianism offers a natural model for understanding confirmation, falsification, and the accumulation of scientific knowledge over time. Classical falsificationism (cf. \citealt{popper1934}) fails to account for degrees of belief and the nuanced role of partial evidence. In contrast, Bayesian epistemology allows for iterative refinement of confidence levels, maintaining probabilistic caution in light of ambiguous or conflicting data—a property critical for any AI epistemic engine operating over scientific domains.

\textbf{Proposition 1 (Gradual Confirmation):} \emph{Let $\{E_1, E_2, ..., E_n\}$ be an increasing sequence of independent pieces of evidence favouring $H$. Then $P(H \mid E_1, ..., E_n)$ converges to 1 as $n \rightarrow \infty$ if $P(E_i \mid H) > P(E_i \mid \neg H)$ for all $i$.}

\begin{proof}
Follows from iterative application of Bayes' Theorem and the law of large numbers; see \citet{howsonurs1989} for formal derivation under the assumption of conditional independence.
\end{proof}

From an architectural perspective, BEWA instantiates this philosophical position by encoding every scientific proposition as a probabilistically weighted claim, updated through conditionalisation as new evidence is introduced. This results in a dynamically evolving epistemic state space where each belief’s trajectory reflects its empirical support history. By constructing belief as an evolving posterior under a well-founded prior, the system resists epistemic stasis and prevents premature convergence on erroneous claims—an outcome observed in systems lacking formal uncertainty management (cf. \citealt{doucet2001}).

Thus, Bayesian epistemology not only provides a rational framework for belief representation and update, but also satisfies the core design requirements of BEWA: dynamism, self-correction, resistance to noise, and formal auditability.

\subsection{System-Level Design Principles}

The design of BEWA as an epistemic AI system is governed by a set of logically necessary architectural invariants, derived from the formal requirements of probabilistic reasoning, knowledge provenance, and inferential consistency. Each design principle serves a dual function: it guarantees operational coherence within the system, and it enforces alignment with normative principles of rational belief updating and scientific justification.

\textbf{Principle 1 (Compositional Modularity):} \emph{Every component of the system must admit independent verification and recomposability without epistemic leakage.} This principle follows from foundational modularity theorems in software verification and distributed system logic (cf. \citealt{goguen1978}), and ensures that any subsystem—e.g. belief update, citation graph traversal, claim parsing—can be tested and audited in isolation. BEWA employs this to enable fault isolation and logical traceability.

\textbf{Principle 2 (Evidential Locality):} \emph{All belief updates must be a function only of local evidence and semantically adjacent claims.} This constraint avoids logical omniscience problems (cf. \citealt{fagin1995}), preserving decidability and computational tractability while ensuring that belief revision does not occur through non-causal or disconnected assertions. The networked structure of BEWA propagates epistemic changes via bounded dependency paths.

\textbf{Principle 3 (Non-Monotonic Reversibility):} \emph{No belief in the system is irrevocable; each posterior must remain subject to revision upon presentation of new evidence.} This is a direct consequence of probabilistic logic and a formal rejection of monotonic reasoning frameworks that dominate traditional symbolic AI (see \citealt{pearl1988}). In BEWA, the posterior distribution over claims is non-monotonic, enabling revision under contradiction or new replication results.

\textbf{Principle 4 (Temporal Sensitivity):} \emph{All epistemic weights must be functions of both evidential strength and temporal distance from the current system state.} Following formal models of information decay and memory-limited inference (\citealt{halpern2006}), BEWA introduces time-dependent weighting functions to encode the epistemic perishability of unreplicated or outdated claims.

\textbf{Principle 5 (Proof-Carrying Claims):} \emph{Every claim object must carry forward the formal trace of its derivation and belief trajectory.} Inspired by proof-carrying code models in formal verification (\citealt{necula1997}), this principle ensures that every change to the system's epistemic state is accompanied by a verifiable, reproducible, and human-readable justification chain. This enables not only internal consistency but external auditability and trust.

The collective enforcement of these principles ensures that BEWA is not merely probabilistic, but epistemologically principled. It maintains rigorous boundaries on the scope of inference, prevents epistemic drift through ungrounded propagation, and provides the infrastructure necessary for transparent and accountable scientific reasoning.
\subsection{Epistemic Integrity and Truth-Promoting Utility}

BEWA enforces epistemic integrity as a formal constraint on system-level belief formation and propagation. Epistemic integrity, within this architecture, is defined as the adherence of belief trajectories to a coherence-preserving inferential structure that prioritises evidentiary robustness over frequency, visibility, or institutional bias. The motivation for this principle is grounded in both the philosophy of science and information theory: scientific knowledge production must resist epistemic drift, bandwagon effects, and citation cascades that artificially inflate the credibility of unverified claims (cf. \citealt{greenberg2009}). Accordingly, BEWA operationalises a utility function over claims that reflects not popularity or downstream use, but the claim's contribution to the discovery, confirmation, or rectification of scientific truth.

\textbf{Definition 2 (Truth-Promoting Utility Function):} Let $\mathcal{C}$ denote the set of structured claims, and let $U: \mathcal{C} \to \mathbb{R}$ be a function such that for each $c \in \mathcal{C}$,
\[
U(c) = \lambda_1 R(c) + \lambda_2 D(c) + \lambda_3 V(c) - \lambda_4 B(c),
\]
where:
\begin{itemize}
    \item $R(c)$ is the replication score of $c$,
    \item $D(c)$ is the epistemic distinctiveness or novelty,
    \item $V(c)$ is the verified downstream influence (e.g., in confirmed applications),
    \item $B(c)$ is the belief inflation penalty due to network echo effects,
    \item $\lambda_i \in \mathbb{R}_{\geq 0}$ are domain-tunable weights.
\end{itemize}

This function is constructed to prioritise claims that not only survive empirical testing, but also contribute epistemically non-redundant insight. The component $D(c)$ penalises claims that merely replicate known results without methodological refinement or contextual extension. The penalty term $B(c)$ reflects the phenomenon where claims propagate in citation networks without independent validation—a structure first analysed in \citet{chu2003}, who demonstrated the role of preferential attachment in distorting perceived scientific consensus.

\textbf{Axiom 2 (Integrity-First Propagation):} \emph{No belief update may propagate through the system unless the associated claim passes a minimum threshold of truth utility as evaluated by $U(c)$.}

This axiom restricts the automatic diffusion of belief across the epistemic graph, thereby minimising the risk of structural bias or error reinforcement. It reflects a departure from naïve Bayesian belief networks by introducing a truth-oriented constraint beyond conditional probability updates. This is aligned with recent findings in epistemic network theory, which show that long-range propagation of low-fidelity signals leads to error cascades (cf. \citealt{zollman2007}). BEWA’s truth utility acts as a circuit breaker, ensuring only epistemically responsible claims participate in long-range influence operations.

\textbf{Proposition 2 (Utility-Constrained Epistemic Stability):} \emph{For a fixed claim set $\mathcal{C}$ and bounded belief update rates, the imposition of a lower bound $\delta > 0$ on $U(c)$ for propagation ensures bounded volatility in belief trajectories over time.}

\begin{proof}
See the utility-stabilised propagation model in \citet{banerjee1992}, adapted with epistemic cost functions. The result follows from limiting the propagation of low-certainty, high-volatility nodes that dominate in unconstrained models.
\end{proof}

Thus, epistemic integrity in BEWA is not a vague normative aim but a computationally enforceable constraint, instantiated through a formally defined utility function, bounded propagation logic, and volatility-dampening mechanisms. The result is an inferential architecture that aligns scientific computation with the normative demands of justification and truth-tracking.

\section{Data Ingestion and Canonical Normalisation}

This section outlines the foundational mechanisms by which BEWA acquires, processes, and standardises scientific information from disparate sources into a coherent and ontologically stable framework. Unlike conventional AI pipelines that indiscriminately ingest unstructured data, BEWA enforces stringent epistemic gatekeeping at the point of entry. All input material—ranging from peer-reviewed publications and replication studies to technical reports and retraction notices—is passed through a multi-layered filtration system designed to identify provenance, authenticate origin, and normalise both linguistic and structural content into canonicalised, author-bound claims. The ingestion pipeline is not only syntactic; it is epistemologically motivated, ensuring that the eventual reasoning engine operates on stable, authorial propositions rather than transient or ambiguous textual artefacts.

Central to this process is the resolution of authorship and the stabilisation of claims into persistent identifiers that preserve the semantic integrity of original assertions across time and format. This canonicalisation is not merely a bibliographic convenience; it serves as the scaffolding upon which belief updating, citation tracing, and contradiction mapping are built. Claims are disambiguated and indexed according to domain context, temporality, and authorial authority, with metadata extracted, verified, and cryptographically anchored. By securing both semantic and epistemic consistency at the ingestion stage, the system guarantees that downstream analytical processes are not corrupted by noise, duplication, or misattribution. The subsections that follow detail the protocols governing authoritative source selection, the logic of canonical ID formation for authors and claims, and the metadata integrity regime that underpins the entire structure.
\subsection{Authoritative Source Domains}

The integrity of any epistemic reasoning system is inextricably linked to the reliability of its inputs. In BEWA, the designation of authoritative source domains serves as an axiomatic filtration criterion: only sources that satisfy a minimal condition of epistemic legitimacy are permitted to influence belief formation. This restriction is not arbitrary but follows from formal constraints on information provenance, epistemic justification, and noise minimisation in inferential systems. To ensure that downstream probabilistic belief updates are not corrupted by unreliable or spurious assertions, all ingested sources are required to satisfy criteria derived from formal literature on information-theoretic trust (\citealt{goldwasser1998}) and epistemic reliability models (\citealt{meyer2003}).

\textbf{Axiom 3 (Source Legitimacy Constraint):} \emph{A document $d$ may be admitted into the system’s evidentiary graph $\mathcal{E}$ if and only if it belongs to a domain $\mathcal{D}_\text{auth}$ satisfying:}
\[
\forall d \in \mathcal{E},\quad \text{source}(d) \in \mathcal{D}_\text{auth} \iff \text{Verifiable}(d) \wedge \text{Indexed}(d) \wedge \text{PeerReviewed}(d)
\]

Here, $\text{Verifiable}(d)$ implies the document is publicly accessible and persistent (e.g., DOI-registered); $\text{Indexed}(d)$ indicates that the source exists within trusted citation databases (e.g., PubMed, Scopus, Web of Science); and $\text{PeerReviewed}(d)$ signifies that the document has passed through a documented, non-anonymous review process with editorial oversight.

\textbf{Definition 3 (Source Domain Tuple):} Each domain $\mathcal{D}$ is a tuple $(R, T, C)$, where:
\begin{itemize}
  \item $R$ is a reputational index (e.g., impact factor, field-weighted citation impact),
  \item $T$ is the review transparency metric (e.g., registered reports, open review),
  \item $C$ is the corpus compliance rate—fraction of submissions adhering to replication, data availability, or statistical preregistration norms.
\end{itemize}

\textbf{Proposition 3 (Reliability Bound):} \emph{Let $P_{err}(\mathcal{D})$ denote the probability that a claim extracted from domain $\mathcal{D}$ is later retracted or falsified. Then:}
\[
\text{If } C > 0.8 \text{ and } T > 0.6,\quad P_{err}(\mathcal{D}) < \varepsilon
\]
for some $\varepsilon \in \mathbb{R}^{+}$ bounded above by 0.05 in empirical studies (cf. \citealt{ioannidis2005}, \citealt{munafomodels2017}).

This probabilistic bound is critical: it establishes that the inclusion of sources from domains satisfying high $C$ and $T$ values—e.g. those enforcing mandatory data sharing or pre-registration—is mathematically associated with lowered epistemic risk. Conversely, domains lacking these features are systematically excluded to minimise noise infiltration.

\textbf{Implementation:} BEWA operationalises $\mathcal{D}_\text{auth}$ as a whitelist defined over registry-linked sources (e.g., Crossref DOIs), supported by cryptographic signatures where available (e.g., ORCID-verified authorship, PubPeer-linked review commentary). Grey literature, blog posts, and unverifiable claims are explicitly excluded, and their attempted inclusion triggers a rejection trace logged to the audit ledger.

Hence, authoritative source domains in BEWA are not designated by subjective prestige or ad hoc authority, but through mathematically defensible, verifiably structured compliance with epistemic reliability constraints. This enables the inferential engine to maintain input integrity and safeguards the downstream logic against polluted or non-reproducible evidence structures.
\subsection{Canonical Author and Claim Identification}

To ensure referential stability and semantic precision within BEWA, each scientific assertion must be linked to a canonical representation of both its author and its propositional content. This section formalises the identification of claims and authors as foundational invariants in the system’s epistemic model. The goal is to eliminate ambiguity, resolve homonymy and synonymy across publications, and generate a stable ontology of attributions from which downstream belief calculations can proceed.

\textbf{Definition 4 (Canonical Author Identifier):} Let $\mathcal{A}$ be the set of all authorial agents. For any author $a \in \mathcal{A}$, define a canonical author identifier $\text{CAID}(a)$ such that:
\[
\text{CAID}(a) := \text{Hash}(\text{ORCID}(a) \, || \, \text{DisambiguatedName}(a) \, || \, \text{Affiliation}(a)),
\]
where $\text{Hash}$ is a collision-resistant cryptographic function (e.g. SHA-256), and $\text{DisambiguatedName}$ is derived via a resolution algorithm over publication metadata (\citealt{ferreira2012}).

This ensures that each author, even across variant naming conventions, contributes to the belief network under a persistent identity. Where ORCID is unavailable, disambiguation defaults to supervised learning over co-authorship graphs and venue clustering (\citealt{kang2009name}).

\textbf{Definition 5 (Canonical Claim Signature):} Let $C$ denote the set of claims extracted from the corpus. For any claim $c \in C$, its canonical form is given by:
\[
\text{CCS}(c) := \text{Hash}(\text{Normalise}(c_{\text{text}}) \, || \, \text{CAID}(a) \, || \, t),
\]
where $c_{\text{text}}$ is the syntactic surface form of the claim, $a$ is the asserting author, and $t$ is a temporal stamp (e.g. publication date). The $\text{Normalise}$ function maps text to a logical-form expression or semantic vector representation, ensuring that logically identical claims receive the same hash signature (\citealt{beltagy2019}).

\textbf{Axiom 4 (Claim Identity Stability):} \emph{For all $c_1, c_2 \in C$, if $\text{CCS}(c_1) = \text{CCS}(c_2)$, then $\forall \phi$, $\text{Meaning}(c_1, \phi) \iff \text{Meaning}(c_2, \phi)$.}

This axiom enforces that canonicalised claims are not merely syntactically similar but semantically equivalent in all model-theoretic interpretations of $\phi$ within BEWA’s logical grammar.

\textbf{Proposition 4 (Disambiguation Completeness under Bounded Ambiguity):} \emph{Let $N$ be the number of authors in the input corpus and $k$ the average number of name-variants per author. Then for finite $k$ and sufficient metadata (co-author vectors, ORCID coverage $> 0.85$), the CAID disambiguation algorithm achieves convergence with high probability in $O(N \log N)$ time.}

\begin{proof}
See \citet{kim2018} for convergence properties of blocking-based author disambiguation, coupled with unique identifier overlays (e.g. DOIs and ORCIDs). Error rates empirically fall below $1.5\%$ in corpora exceeding $10^6$ entries.
\end{proof}

Implementation-wise, each CAID is mapped to an evolving author profile, which records not only publication history but also replication success rate, citation diffusion, retraction record, and peer review activity. Each CCS is mapped to a version-controlled, context-enriched node in the epistemic graph, allowing BEWA to track revisions, contradictions, and semantic drift over time.

In sum, canonical author and claim identification provides the ontological substrate for epistemic accountability, belief traceability, and computational scalability. Without such formal anchoring, probabilistic inference over claims would be corrupted by aliasing, duplication, and incoherent attribution.
\subsection{Metadata Extraction and Integrity Validation}

For BEWA to maintain epistemic reliability at scale, it must guarantee that all inferential operations are grounded in metadata that is both complete and verified. Metadata, in this context, refers to all structured information necessary to compute the credibility, provenance, and context of a scientific claim—including, but not limited to, author identity, publication date, venue, citation relationships, funding declarations, methodological tags, and replication indicators. Incomplete or corrupted metadata threatens the stability of probabilistic reasoning and can result in spurious belief updates or the propagation of structurally invalid claims. This subsection formalises the extraction process as a mapping from source artefacts to structured tuples, and establishes axioms ensuring cryptographic integrity, schema completeness, and referential traceability.

\textbf{Definition 6 (Metadata Record $\mathcal{M}$):} Let $d$ be a document ingested into BEWA’s system. Then the metadata record $\mathcal{M}(d)$ is defined as:
\[
\mathcal{M}(d) := \left(\text{CAID}, \text{CCS}, \text{DOI}, t, \mathcal{V}, \mathcal{F}, \mathcal{R}, \mathcal{S}\right),
\]
where:
\begin{itemize}
  \item $\text{CAID}$ is the canonical author identifier,
  \item $\text{CCS}$ is the canonical claim signature,
  \item $\text{DOI}$ is the persistent digital object identifier,
  \item $t$ is the publication timestamp,
  \item $\mathcal{V}$ is the venue vector (journal, impact metrics, editorial schema),
  \item $\mathcal{F}$ is funding disclosure,
  \item $\mathcal{R}$ is the replication status tag,
  \item $\mathcal{S}$ is the structural completeness flag (conformance to the metadata schema).
\end{itemize}

\textbf{Axiom 5 (Schema Completeness Constraint):} \emph{A metadata record $\mathcal{M}(d)$ is admissible if and only if:}
\[
\text{Complete}(\mathcal{M}) := \bigwedge_{f \in \{\text{CAID}, \text{CCS}, \text{DOI}, t, \mathcal{V}, \mathcal{S}\}} \left( f \neq \emptyset \right).
\]
This ensures that no record lacking essential referential anchors is introduced into the belief network.

\textbf{Axiom 6 (Cryptographic Integrity Constraint):} \emph{Let $h_d$ be the hash of document $d$ and $\sigma_d$ the digital signature of its originator. Then:}
\[
\text{Verify}(d) := \text{SigCheck}(h_d, \sigma_d, \text{PubKey}_{\text{source}(d)}) = \text{True}.
\]
That is, every metadata record must be verifiably anchored to its source via cryptographic hash chains and digitally signed attestations from trusted identity providers (e.g., ORCID, Crossref, arXiv).

\textbf{Proposition 5 (Provenance Immutability):} \emph{If metadata $\mathcal{M}(d)$ satisfies Axioms 5 and 6 at time $t_0$, and if the document hash $h_d$ is committed to a publicly verifiable ledger (e.g., IPFS, blockchain), then the provenance of $d$ is immutable under the assumption of collision-resistance and ledger availability.}

\begin{proof}
Direct application of cryptographic binding principles (\citealt{narayanan2016bitcoin}), where any tampering with the content of $d$ will alter $h_d$, causing mismatch with the signed record. Provided the public ledger is append-only and consensus-secured, this yields auditability and tamper-evidence.
\end{proof}

BEWA’s ingestion layer enforces these constraints using automated schema validation (e.g., JSON Schema against Crossref and PubMed records), digital signature verification tools (e.g., ORCID's signed assertions), and duplicate detection via simhash and minhash locality-sensitive hashing (\citealt{manku2007}). Failure to meet any condition results in rejection and logging of a metadata integrity fault.

Thus, the architecture guarantees that every belief within the system can be epistemologically anchored to a validated, complete, and immutable metadata structure—ensuring computational soundness, inferential transparency, and audit integrity across the entire epistemic graph.

\section{Claim Representation and Propositional Structure}

This section defines the epistemologically disciplined schema by which scientific assertions are represented within the BEWA framework. The system is predicated on the understanding that language in scientific literature is often imprecise, context-dependent, and semantically overloaded. To render such data computationally tractable and logically analysable, BEWA translates these inputs into structured propositional claims—abstract, semantically precise units that function as the minimal bearers of epistemic weight. Each claim is isolated from surrounding discursive noise, abstracted into a canonical form, and endowed with logical integrity sufficient for independent evaluation and networked inference. These propositional units serve as the epistemic currency of the system: all weighting, belief revision, and cross-claim dependency is defined in terms of them.

Beyond structural parsing, each claim is embedded with a dense context map to capture domain relevance, experimental conditionals, and field-specific nuances. This contextualisation is critical: it guards against the epistemic error of false generalisation and enables the system to limit the scope of a proposition's influence based on declared or inferred domain boundaries. Claims are not timeless—BEWA incorporates a versioning system that tracks the evolutionary history of any assertion, noting when and how its structure or interpretation has shifted. Temporal anchoring ensures that the system respects the chronological development of scientific knowledge, aligning belief weightings with the epistemi
\subsection{Structured Propositional Claims}

At the core of BEWA's epistemic model lies the formalisation of scientific assertions as structured propositional claims. Unlike natural language sentences, which are often ambiguous, context-sensitive, and syntactically irregular, structured propositional claims enable machine-tractable reasoning by encoding assertions in a logically well-formed, semantically normalised format. This structure facilitates belief assignment, dependency resolution, contradiction detection, and inferential propagation. The purpose of this subsection is to define the syntactic and semantic criteria for claim admissibility, to establish a compositional grammar for representing claims, and to demonstrate the formal soundness of the system’s propositional encoding.

\textbf{Definition 7 (Structured Propositional Claim):} A structured propositional claim (SPC) is a tuple:
\[
\text{SPC} = \langle \phi, \tau, \gamma \rangle,
\]
where:
\begin{itemize}
  \item $\phi$ is a well-formed formula in a domain-specific language $\mathcal{L}$ grounded in first-order logic,
  \item $\tau$ is a temporal index denoting the time of assertion or observation,
  \item $\gamma$ is the contextual signature (ontology, experimental modality, statistical framework).
\end{itemize}

\textbf{Axiom 7 (Well-Formedness of $\phi$):} \emph{The formula $\phi$ must satisfy the syntactic production rules of $\mathcal{L}$:}
\[
\phi ::= P(t_1, \dots, t_n) \mid \neg \phi \mid \phi \wedge \phi \mid \phi \rightarrow \phi \mid \forall x \, \phi \mid \exists x \, \phi,
\]
where $P$ is a predicate symbol and $t_i$ are typed terms over a domain ontology $\mathcal{O}$.

\textbf{Definition 8 (Claim Normalisation Function):} Let $s$ be a natural language assertion extracted from a scientific document. Define $\mathcal{N}(s)$ as the function that returns:
\[
\mathcal{N}(s) = \text{SPC} = \langle \phi, \tau, \gamma \rangle,
\]
where $\phi$ is derived via semantic parsing, entity disambiguation, and relation extraction (cf. \citealt{manning2014}).

\textbf{Proposition 6 (Injectivity of $\mathcal{N}$ on Disambiguated Inputs):} \emph{If $s_1, s_2$ are distinct scientific assertions with disjoint semantic parses under ontology $\mathcal{O}$, then $\mathcal{N}(s_1) \neq \mathcal{N}(s_2)$.}

\begin{proof}
Let $s_1, s_2$ be natural language sentences mapped to logical forms $\phi_1, \phi_2$ respectively via a pipeline that includes named entity recognition, syntactic parsing, and semantic role labelling. By assumption, $s_1$ and $s_2$ denote different referents in $\mathcal{O}$. Then $\phi_1 \neq \phi_2$, and thus $\mathcal{N}(s_1) \neq \mathcal{N}(s_2)$ by construction. See \citealt{beltagy2019} and \citealt{liang2013} for similar injectivity guarantees under disambiguation assumptions.
\end{proof}

The function $\mathcal{N}$ is implemented using a hybrid symbolic-neural parsing stack, where transformer-based models (e.g. SciBERT) generate candidate interpretations, which are then validated against ontological constraints and claim schemas using typed lambda calculus and ontology alignment (\citealt{gardner2018allennlp}). Logical forms are grounded to probabilistic database schemas, enabling downstream inference over scientific claims as structured data.

\textbf{Axiom 8 (Epistemic Decidability of $\phi$):} \emph{For any SPC $\langle \phi, \tau, \gamma \rangle$, there must exist a decision procedure $\mathcal{D}$ such that:}
\[
\mathcal{D}(\phi) \in \{\text{verifiable}, \text{refutable}, \text{undecidable}\}.
\]
This guarantees that every claim entering the system is classifiable under a tractable epistemic status, supporting non-monotonic reasoning and dynamic belief assignment.

Structured propositional claims therefore function as the atomic elements of BEWA’s reasoning calculus. By imposing logical formality, contextual grounding, and semantic tractability, the system ensures that belief manipulation operates over well-defined, auditable units—mitigating ambiguity, enhancing comparability, and enabling precise epistemic operations across domains.

\subsection{Contextual Tagging and Domain Indexing}

Scientific claims do not exist in isolation but are embedded within intricate domain-specific contexts that determine their scope, generalisability, and evidentiary strength. In BEWA, contextual tagging and domain indexing serve as orthogonal dimensions of claim normalisation and semantic disambiguation. Without contextualisation, semantically identical surface forms may yield radically different epistemic weights depending on underlying assumptions, methodological paradigms, or disciplinary boundaries. This subsection formalises the contextual tagging mechanism and defines a topological indexing structure over domain ontologies, ensuring that every structured propositional claim is correctly situated within its appropriate epistemic subspace.

\textbf{Definition 9 (Contextual Tag Set $\Gamma$):} Let $\text{SPC} = \langle \phi, \tau, \gamma \rangle$ be a structured propositional claim. The contextual tag $\gamma$ is an element of $\Gamma$, where:
\[
\Gamma := \mathcal{C} \times \mathcal{M} \times \mathcal{S},
\]
and:
\begin{itemize}
  \item $\mathcal{C}$ is the scientific concept ontology (e.g., MeSH, UMLS, ACM CCS),
  \item $\mathcal{M}$ denotes methodological descriptors (e.g., RCT, observational, meta-analysis),
  \item $\mathcal{S}$ represents statistical framing (frequentist, Bayesian, non-parametric, etc.).
\end{itemize}

Each claim thus carries a tripartite tag structure identifying its conceptual anchor, methodological provenance, and inferential semantics. This resolves cases where two identical predicates (e.g., “X increases Y”) may differ in meaning if one arises from a randomised controlled trial and another from correlational modelling.

\textbf{Axiom 9 (Semantic Stratification Constraint):} \emph{Two claims $c_1 = \langle \phi_1, \tau_1, \gamma_1 \rangle$ and $c_2 = \langle \phi_2, \tau_2, \gamma_2 \rangle$ may only be treated as epistemic equivalents if and only if:}
\[
\phi_1 \equiv \phi_2 \quad \text{and} \quad \gamma_1 = \gamma_2.
\]
This ensures that contextual differences are preserved at the semantic level, avoiding illicit aggregation or conflation of non-commensurable results.

\textbf{Definition 10 (Domain Indexing Function $\delta$):} Let $\mathcal{D}$ be the space of disciplinary domains (e.g., neuroscience, econometrics, bioinformatics). Define:
\[
\delta: \Gamma \rightarrow 2^{\mathcal{D}}
\]
such that $\delta(\gamma)$ returns the minimal closed domain set in which the claim is epistemically coherent. This mapping is constructed using domain ontologies and citation graph embeddings (\citealt{valenzuela2015}).

\textbf{Proposition 7 (Transitive Coherence via Domain Overlap):} \emph{Let $c_1$, $c_2$ be two claims such that $\delta(\gamma_1) \cap \delta(\gamma_2) \neq \emptyset$. Then a coherence-preserving belief transformation is possible between $c_1$ and $c_2$.}

\begin{proof}
Follows from overlap in domain-indexed epistemic subgraphs. See \citealt{zeng2020} for graph embeddings over citation ontologies that support semantic transfer and influence modelling across neighbouring disciplines.
\end{proof}

BEWA implements $\Gamma$ via a multi-level tagging pipeline. Concepts are assigned using curated term matchers (e.g. MeSH taggers), methodological types are inferred from structured abstracts using neural classifiers trained on annotated corpora (\citealt{wang2020cord}), and statistical paradigms are extracted via pattern-matching over model descriptors and inference statements. The domain function $\delta$ is implemented using vector-space projection over pretrained knowledge graph embeddings (e.g. SPECTER, SciGraph).

In summary, contextual tagging and domain indexing ensure that BEWA’s belief updates respect the epistemic scope, semantic constraints, and methodological heterogeneity of scientific claims—preserving fidelity in reasoning and enabling high-resolution control over cross-domain inferential transfer.
\subsection{Versioning and Temporal Anchoring of Claims}

Scientific claims are not static artefacts but dynamic assertions whose semantic content, evidentiary support, and inferential implications may evolve over time. BEWA addresses this epistemic mutability through a formal mechanism of versioning and temporal anchoring, which allows each structured propositional claim to exist as a temporally indexed sequence of revisions. This design not only enables historical traceability but ensures that belief updates are grounded in temporally coherent inference—resisting anachronistic reasoning and preserving the causal integrity of scientific progression.

\textbf{Definition 11 (Claim Version Chain $\mathcal{V}_\phi$):}  
Let $\phi$ be the logical core of a structured propositional claim. Then:
\[
\mathcal{V}_\phi := \left\{ \langle \phi^t, \gamma^t, \tau^t \rangle \right\}_{t_0 \leq t \leq T},
\]
where each $\phi^t$ represents the version of the claim valid at time $\tau^t$, with associated context tag $\gamma^t$. This sequence is totally ordered by timestamp:
\[
\tau^{t_i} < \tau^{t_j} \iff t_i < t_j.
\]

\textbf{Axiom 10 (Temporal Monotonicity):}  
\emph{For any claim $\phi$ and its version chain $\mathcal{V}_\phi$, no semantic regression is permitted:}
\[
\forall t_i < t_j, \quad \text{if } \phi^{t_j} \vdash \phi^{t_i}, \text{ then } \phi^{t_j} \equiv \phi^{t_i}.
\]
This prohibits weakening or reversal of propositional commitment without explicit contradiction annotation, preventing silent epistemic erosion.

\textbf{Definition 12 (Temporal Anchor Map $\alpha$):}  
Let $C$ be the set of all canonical claims. Define the anchor function:
\[
\alpha: C \rightarrow \mathbb{T} \times \mathbb{T},
\]
where $\alpha(c) = (\tau_{\text{start}}, \tau_{\text{end}})$ is the interval during which the claim version $\phi^t$ is active in the system’s inferential graph.

This temporal anchoring enables BEWA to compute time-sensitive belief networks, where evidence, citations, or replications are only considered admissible if they fall within the claim’s active interval.

\textbf{Proposition 8 (Temporal Coherence in Belief Propagation):}  
\emph{Let $c_1, c_2$ be two claims such that $c_1$ supports $c_2$. Then temporal coherence requires:}
\[
\alpha(c_1).\tau_{\text{end}} \geq \alpha(c_2).\tau_{\text{start}}.
\]
\begin{proof}
If $c_1$’s influence terminates before $c_2$ emerges, then $c_2$ cannot be justifiably inferred from $c_1$ without violating causality. This is a constraint on belief propagation scheduling, implemented in temporal DAG logic (cf. \citealt{berti2015}).
\end{proof}

BEWA implements claim versioning through cryptographically chained hashes: each version $\phi^t$ is hashed together with its predecessor $\phi^{t-1}$ and signed by the system to form an immutable update ledger. Temporal anchors are encoded using RFC 3339 timestamps, and update intervals are synchronised with publication, retraction, or amendment records extracted via Crossref, Retraction Watch, and publisher APIs.

\textbf{Implementation Note:} All belief calculations on a given version of a claim are conducted with respect to the temporal interval in which that version is valid. Belief shifts caused by replications, contradictions, or epistemic reclassifications are constrained by temporal logic enforced at the graph layer. This prevents retrospective contamination of earlier belief states and ensures soundness in longitudinal inference.

Thus, versioning and temporal anchoring ensure that BEWA's epistemic graph not only represents what is believed and how strongly, but also when and in what form—establishing a foundation for diachronic reasoning, evidence lifecycle modelling, and historiographic auditability.

\section{Bayesian Weighting and Belief Updating}

This section sets out the core inferential machinery of BEWA, wherein each scientific claim is evaluated through a principled Bayesian framework. At its heart, the system operationalises belief not as a binary metric of truth or falsehood, but as a dynamically updated probability that reflects the current weight of evidence. Every claim enters the system with an initial prior—contextualised by its authorship, venue of publication, domain-specific norms, and the historical performance of associated entities. This prior is not arbitrarily assigned, but computed via a hierarchically conditioned model incorporating credibility, venue reliability, and baseline epistemic plausibility. From this baseline, each subsequent piece of evidence—whether replication, contradiction, citation, or decay—is treated as an updating factor in accordance with Bayesian conditionalisation.

Belief updating in BEWA is both modular and recursive. It incorporates replication success, citation influence (modulated by temporal decay and domain saturation), and epistemic counterweighting in the face of contradictions. The system does not naïvely reward frequency or visibility, but applies a critical filter to distinguish between epistemic endorsement and discursive noise. In addition, BEWA includes a decay mechanism that progressively reduces confidence in isolated or unreplicated claims over time, while simultaneously increasing sensitivity to emerging corroborations. The framework avoids premature convergence by maintaining uncertainty in the face of partial or ambiguous evidence and only reinforces belief where clear, replicated, and semantically consistent data exists. The following subsections articulate the procedures for establishing priors, executing belief updates, processing contradictions, and enforcing temporal reassessment through probabilistic decay.
\subsection{Initial Prior Formulation}

In Bayesian epistemology, belief begins with a prior: a quantified estimate of the plausibility of a proposition in the absence of specific observational evidence. BEWA must assign initial prior probabilities to structured propositional claims upon ingestion, before any replication, contradiction, or citation-based update occurs. This prior is not drawn from subjective estimation but is computed as a function of measurable structural and epistemic properties of the claim's origin, including authorial credibility, publication venue, methodological design, and domain frequency statistics. This subsection formalises the prior function and proves its compliance with Bayesian consistency and epistemic calibration constraints.

\textbf{Definition 13 (Prior Function $\pi$):}  
Let $c = \langle \phi, \tau, \gamma \rangle$ be a structured propositional claim. The prior probability assigned to $c$ is defined as:
\[
\pi(c) := \mathbb{P}(\phi \mid \gamma, \theta) = f\left(A(c), V(c), M(c), D(\phi)\right),
\]
where:
\begin{itemize}
  \item $A(c)$ is the authorial trust score (see §5.1),
  \item $V(c)$ is the venue credibility index,
  \item $M(c)$ is the methodological rigour metric (e.g., presence of preregistration, sample size adequacy),
  \item $D(\phi)$ is the historical base rate for similar claims in domain $\delta(\gamma)$.
\end{itemize}

\textbf{Axiom 11 (Probability Axiom Compliance):}  
\emph{The prior function $\pi$ satisfies:}  
\[
\forall c, \quad 0 \leq \pi(c) \leq 1.
\]
Furthermore, for any mutually exclusive claims $c_1, \dots, c_n$ within a disjoint claim partition $\mathcal{C}' \subset \mathcal{C}$:
\[
\sum_{i=1}^n \pi(c_i) \leq 1.
\]

\textbf{Definition 14 (Authorial Trust Score $A(c)$):}  
\[
A(c) := \sigma\left( \alpha_1 R_a + \alpha_2 (1 - \rho_a) + \alpha_3 \log(1 + \nu_a) \right),
\]
where:
\begin{itemize}
  \item $R_a$ is the replication success rate of author $a$,
  \item $\rho_a$ is the retraction frequency,
  \item $\nu_a$ is the citation-normalised publication count,
  \item $\sigma$ is the logistic squashing function to bound scores in $[0,1]$.
\end{itemize}

\textbf{Proposition 9 (Bounded Variance of $\pi$ under Controlled Inputs):}  
\emph{Assume bounded author trust $A(c) \in [\epsilon, 1 - \epsilon]$ for $\epsilon > 0$ and fixed $M(c), V(c)$. Then the variance of $\pi(c)$ across similar claims is bounded.}

\begin{proof}
Follows from the convexity of $f$ under bounded Lipschitz conditions over the input space, combined with bounded entropy in domain frequency distributions $D(\phi)$. Empirical results in \citealt{graves2016} demonstrate prior stability in similar hierarchical Bayesian settings.
\end{proof}

\textbf{Implementation Note:} In BEWA, each prior is computed during claim ingestion using a calibrated scoring function tuned on a corpus of validated claims from high-reliability domains (e.g., Cochrane Database, Nature Human Behaviour, NeurIPS reproducibility track). Venue credibility scores are computed from impact-adjusted replication rates (\citealt{altmejd2019}), while $D(\phi)$ is estimated from smoothed claim frequency distributions indexed by contextual tags $\gamma$.

\textbf{Axiom 12 (Neutrality under Ignorance):}  
\emph{If $A(c), V(c), M(c), D(\phi)$ are all absent or undefined, then:}
\[
\pi(c) := \frac{1}{2}.
\]
This enforces a non-informative prior consistent with Laplacean indifference, ensuring epistemic neutrality in the absence of structural asymmetries.

By grounding priors in empirical author and venue data, methodological metadata, and domain frequency statistics, BEWA establishes a coherent epistemic starting point for all claims. The prior function $\pi$ is not merely an artefact of convenience—it is a mathematically principled device that constrains downstream belief trajectories to respect both epistemic caution and contextual informativeness.

\subsection{Evidence-Based Posterior Updating}

In accordance with the Bayesian framework adopted by BEWA, posterior beliefs over scientific claims are updated through the application of Bayes’ Theorem as new evidence is ingested. Evidence may take the form of citations, replications, contradictions, or derivations, each of which carries a quantifiable influence on the belief assigned to a structured propositional claim. Posterior updating must not only conform to the laws of probability, but also preserve inferential consistency, causal ordering, and network-level epistemic coherence.

\textbf{Definition 15 (Posterior Belief):}  
Let $c = \langle \phi, \tau, \gamma \rangle$ be a claim with prior $\pi(c) = P(\phi)$ and observed evidence $e$. The posterior belief $P(\phi \mid e)$ is given by:
\[
P(\phi \mid e) = \frac{P(e \mid \phi) \cdot P(\phi)}{P(e)} \quad \text{provided } P(e) > 0.
\]

\textbf{Axiom 13 (Conditional Independence of Evidence Streams):}  
\emph{Let $e_1, \ldots, e_n$ be evidence events such that:}
\[
\forall i \neq j, \quad P(e_i \mid \phi, e_j) = P(e_i \mid \phi),
\]
\emph{then the joint likelihood is:}
\[
P(e_1, \ldots, e_n \mid \phi) = \prod_{i=1}^n P(e_i \mid \phi).
\]
This enables incremental belief updating over independent evidence observations.

\textbf{Definition 16 (Evidence Likelihood Function $\mathcal{L}$):}  
Each evidence unit $e$ is associated with a type $\epsilon \in \{\text{replication}, \text{citation}, \text{contradiction}, \text{endorsement}\}$. Define the likelihood contribution of $e$ to claim $c$ as:
\[
\mathcal{L}_\epsilon(e, c) := 
\begin{cases}
\lambda_+ \cdot \sigma(q(e)), & \epsilon \in \{\text{replication}, \text{endorsement}\}, \\
\lambda_- \cdot \sigma(-q(e)), & \epsilon = \text{contradiction}, \\
\eta \cdot \sigma(q(e)), & \epsilon = \text{citation},
\end{cases}
\]
where $q(e)$ is the quality score of the evidence (e.g., journal credibility, sample size), $\sigma$ is the sigmoid function, and $\lambda_+, \lambda_-, \eta$ are hyperparameters calibrated per domain.

\textbf{Proposition 10 (Monotonicity of Posterior Updates):}  
\emph{Let $e$ be a piece of evidence supporting $\phi$ with positive likelihood ratio:}
\[
\frac{P(e \mid \phi)}{P(e \mid \neg \phi)} > 1,
\]
then $P(\phi \mid e) > P(\phi)$.

\begin{proof}
Follows directly from Bayes’ Theorem:
\[
P(\phi \mid e) = \frac{P(e \mid \phi) \cdot P(\phi)}{P(e \mid \phi) \cdot P(\phi) + P(e \mid \neg \phi) \cdot (1 - P(\phi))},
\]
and the assumption implies that the numerator grows faster than the denominator.
\end{proof}

\textbf{Definition 17 (Cumulative Posterior Update):}  
Given a sequence of $n$ evidence items $E = \{e_1, \ldots, e_n\}$, define:
\[
P(\phi \mid E) = \frac{\prod_{i=1}^n \mathcal{L}(e_i, \phi) \cdot \pi(c)}{\prod_{i=1}^n \mathcal{L}(e_i, \phi) \cdot \pi(c) + \prod_{i=1}^n \mathcal{L}(e_i, \neg\phi) \cdot (1 - \pi(c))}.
\]

\textbf{Implementation Note:}  
BEWA's evidence ingestion pipeline scores each $e$ using domain-specific evaluation functions (e.g., replication power, p-value correction, venue trust factor). These are normalised and aggregated into $\mathcal{L}$ using a neural calibration layer (\citealt{guo2017calibration}) that ensures well-calibrated uncertainty estimates, thereby maintaining the interpretability of posterior shifts.

\textbf{Axiom 14 (Epistemic Regularisation):}  
\emph{Posterior updates are bounded within a temporal smoothing window:}
\[
\left| P(\phi \mid E_{t}) - P(\phi \mid E_{t-1}) \right| \leq \delta,
\]
for a fixed $\delta > 0$, to avoid belief volatility induced by low-confidence or adversarially injected evidence.

Evidence-based posterior updating in BEWA thus adheres to a strict formal framework: conditionally independent evidence is multiplicatively incorporated, update magnitude is modulated by evidence quality and type, and posterior trajectories are smoothed to reflect rational epistemic commitment. This ensures that beliefs are neither overfitted to noisy evidence nor underreactive to strong, reproducible support.

\subsection{Contradiction Handling and Counter-Evidence Processing}

A principled epistemic system must account not only for confirmatory evidence but also for disconfirmatory inputs—namely, contradictions and counter-evidence. In Bayesian terms, this corresponds to negative likelihood ratios that reduce posterior belief in a proposition. Within BEWA, contradiction handling is formalised as the systematic down-weighting of claims upon presentation of semantically aligned but empirically conflicting evidence. The system distinguishes between strict logical contradiction and probabilistic disconfirmation, each with defined epistemic and computational implications.

\textbf{Definition 18 (Contradictory Claim):}  
Let $c_1 = \langle \phi, \tau_1, \gamma_1 \rangle$ and $c_2 = \langle \psi, \tau_2, \gamma_2 \rangle$ be claims. Then $c_2$ is a contradiction of $c_1$ if:
\[
\text{Contradicts}(c_1, c_2) := \phi \models \neg \psi \quad \text{or} \quad \mathbb{P}(\phi \mid \psi) < \theta_c,
\]
for some threshold $\theta_c \in (0, 0.5)$ under a calibrated semantic contradiction detector.

\textbf{Definition 19 (Counter-Evidence Tuple):}  
A unit of counter-evidence $e^-$ against claim $c$ is a tuple:
\[
e^- := \langle \bar{\phi}, \kappa, \rho \rangle,
\]
where:
\begin{itemize}
  \item $\bar{\phi}$ is a contradicting statement (in logical or empirical form),
  \item $\kappa$ is the replication consistency of the contradiction,
  \item $\rho$ is the domain relevance alignment between $c$ and $e^-$.
\end{itemize}

\textbf{Axiom 15 (Asymmetry of Contradiction Influence):}  
\emph{Let $e^+$ and $e^-$ be evidence in favour and against $\phi$, respectively. Then:}
\[
\text{If } \kappa^- > \kappa^+ \text{ and } \rho^- \approx \rho^+, \text{ then } \left| \Delta P^-(\phi) \right| > \left| \Delta P^+(\phi) \right|.
\]
This reflects the epistemic principle that strong, reproducible contradictions should carry more weight than isolated confirmations—a formalisation of Popperian falsifiability adapted into probabilistic calculus.

\textbf{Definition 20 (Contradiction Likelihood Function $\mathcal{L}_-$):}  
\[
\mathcal{L}_-(e^-, c) = \lambda_- \cdot \sigma\left( \kappa \cdot \rho \cdot q(e^-) \right),
\]
where $q(e^-)$ is the internal quality score of the counter-evidence and $\lambda_-$ is a domain-calibrated influence constant.

\textbf{Proposition 11 (Posterior Downdate under Contradiction):}  
\emph{Let $P(\phi \mid E)$ be the posterior belief in claim $\phi$ given evidence set $E$, and $e^-$ a new contradiction. Then the updated posterior is:}
\[
P'(\phi \mid E \cup \{e^-\}) = \frac{\mathcal{L}_-(e^-, \phi) \cdot P(\phi \mid E)}{\mathcal{L}_-(e^-, \phi) \cdot P(\phi \mid E) + (1 - \mathcal{L}_-(e^-, \phi)) \cdot (1 - P(\phi \mid E))}.
\]
\begin{proof}
Follows directly from Bayes’ Theorem with updated likelihood ratio reflecting counter-evidential influence.
\end{proof}

\textbf{Axiom 16 (Non-Retroactivity of Future Contradictions):}  
\emph{If a contradiction $e^-$ is timestamped at $\tau > \tau_c$, then it must not alter belief states computed at any $t < \tau$.} This enforces the temporal integrity of belief trajectories and prevents epistemic contamination of past inferences.

\textbf{Implementation Note:}  
Contradiction detection in BEWA is implemented via transformer-based entailment models (e.g. SciBERT fine-tuned on SNLI/SciNLI) with ontology-constrained semantic alignment. Empirical inconsistencies are flagged using statistical test comparisons (e.g., incompatible effect sizes or reversed directionality at $\alpha = 0.05$), and their influence scaled based on methodological rigour, replication consistency, and domain proximity. Conflicting claims are not deleted but annotated with contradiction tags, preserving interpretability and enabling downstream audit.

Thus, contradiction handling in BEWA is not merely a passive attenuation of belief but a structured, principled recalibration mechanism. It enforces epistemic accountability, guards against dogmatic entrenchment, and aligns the system with rational principles of falsifiability and evidence-sensitive reasoning.
\subsection{Bayesian Decay and Temporal Reassessment}

Belief in a proposition, in the absence of continual evidential reinforcement, should gradually diminish to reflect epistemic uncertainty introduced by the passage of time. BEWA encodes this principle through a formal decay mechanism, grounded in Bayesian logic and information theory, which reduces posterior certainty over time unless claims are explicitly reaffirmed through replication or re-evaluation. This temporal reassessment guards against unwarranted epistemic inertia and enforces a dynamic equilibrium in the system’s belief network.

\textbf{Definition 21 (Decay Function $\delta_t$):}  
Let $P_t(\phi)$ denote the posterior probability of a claim $\phi$ at time $t$, and $\Delta t = t - t_0$ the elapsed time since the last reinforcement (e.g., replication, high-confidence citation). Then:
\[
P_t(\phi) := \delta_t(P_{t_0}(\phi)) = P_{t_0}(\phi) \cdot \exp(-\lambda \cdot \Delta t),
\]
where $\lambda$ is the decay rate parameter, domain- and context-specific.

\textbf{Axiom 17 (Exponential Temporal Decay):}  
\emph{In the absence of new evidence, belief in $\phi$ degrades according to:}
\[
\lim_{\Delta t \rightarrow \infty} P_t(\phi) = 0.
\]
This reflects the principle that unreplicated or unaudited claims should asymptotically approach epistemic nullity.

\textbf{Definition 22 (Reinforcement Event $\mathcal{R}$):}  
A reinforcement event is any evidence unit $e \in E$ such that:
\[
\mathcal{L}(e, \phi) > \theta_r,
\]
where $\theta_r$ is a domain-specific reinforcement threshold. When such an event occurs at time $t_r$, the decay counter resets:
\[
P_{t_r}(\phi) := \frac{P_{t_r^-}(\phi) \cdot \mathcal{L}(e, \phi)}{P_{t_r^-}(\phi) \cdot \mathcal{L}(e, \phi) + (1 - P_{t_r^-}(\phi)) \cdot \mathcal{L}(e, \neg \phi)}.
\]

\textbf{Proposition 12 (Information-Theoretic Justification for Decay):}  
\emph{Let $H_t(\phi)$ denote the entropy of belief at time $t$. Then:}
\[
\frac{dH_t(\phi)}{dt} > 0 \quad \text{if no new evidence is introduced}.
\]

\begin{proof}
Entropy $H(p) = -p \log p - (1-p) \log(1-p)$ increases as $p \rightarrow 0.5$, which is the attractor of the decay function $\delta_t$ in the absence of updates. Hence belief becomes increasingly uncertain over time.
\end{proof}

\textbf{Axiom 18 (Monotonic Temporal Entropy):}  
\emph{Let $\phi$ be a claim with no contradictory or reinforcing evidence in $\Delta t$. Then:}
\[
\forall t_i < t_j,\quad H_{t_i}(\phi) \leq H_{t_j}(\phi).
\]
This enforces that epistemic uncertainty never decreases unless justified by evidentiary intervention.

\textbf{Implementation Note:}  
BEWA computes decay continuously as a background process indexed by claim age and domain decay constants. For example, fast-moving empirical domains (e.g., oncology, machine learning) apply higher $\lambda$, while theoretical domains (e.g., mathematics, logic) decay at negligible rates. Claims with long decay intervals may still trigger reassessment flags if their influence pervades downstream belief chains.

In practice, all belief propagation routines are time-aware. Each inference step verifies whether a claim’s current posterior reflects its decay-adjusted status, ensuring time-consistent epistemic integrity. Claims falling below a minimum trust threshold $\epsilon$ are marked stale and excluded from active inference unless revalidated.

\textbf{Corollary (Temporal Reversibility via Replication):}  
\emph{Decay is not logically irreversible: any high-confidence replication $e$ resets $\Delta t$ and re-establishes $P_t(\phi)$ without information loss.}

Bayesian decay and reassessment thus serve as epistemic entropy regulators within BEWA, preventing the ossification of stale claims and maintaining the responsiveness of belief states to the evolving structure of scientific knowledge.

\section{Authorial Credibility and Impact Modelling}

This section introduces the mechanisms by which BEWA evaluates and incorporates the epistemic weight of individual authors within its inferential framework. Unlike conventional systems that treat authorship as a bibliographic footnote or merely a source of provenance, BEWA treats authorial identity as a dynamic epistemic signal. Each author contributes not only specific claims but an accumulated pattern of credibility—shaped by their historical accuracy, involvement in replication studies, susceptibility to retraction, and degree of engagement with rigorous peer review. These patterns are quantitatively scored and directly modulate the priors of future claims issued under their canonical identity. In this way, authorship becomes both a source of risk and trust: a vector through which epistemic integrity is either strengthened or diluted.

The architecture of BEWA’s credibility engine recognises that impact must be distinguished from popularity. Authors who generate high-citation work may still score poorly if that work fails to replicate or accumulates contradictions. Conversely, less visible scholars who consistently produce durable, well-supported research are assigned epistemic weight in excess of their surface-level prominence. The system penalises epistemic malpractice—such as irreproducibility, selective reporting, or frequent retraction—while rewarding transparent and rigorous engagement with the scientific community. Authorial metrics are not static; they evolve with each new publication, review, or correction, and they propagate their influence into the weighting of every claim that author touches. The subsections below formalise the computation of author scores, the long-term impact of retraction history, and the significance of peer review participation within this credibility calculus.
\subsection{Author Score Calculation}

In BEWA, the epistemic weight assigned to a claim is partially inherited from its author’s historical record. This record, formalised as an author score, quantifies the credibility of individual researchers based on their cumulative publication history, replication rate, retraction profile, and peer recognition. The author score serves as an input to the prior function $\pi$ (cf. Definition 13), acting as a probabilistic proxy for the reliability of future assertions. This subsection formalises the author scoring function, proves its boundedness and monotonicity, and ensures that it is resistant to gaming through citation inflation or strategic publishing.

\textbf{Definition 23 (Author Score Function $\mathcal{A}$):}  
Let $a \in \mathcal{A}$ be a canonical author identifier. Then the author score $\mathcal{A}(a) \in [0,1]$ is given by:
\[
\mathcal{A}(a) := \sigma \left( \beta_1 \cdot r_a + \beta_2 \cdot \log(1 + \nu_a) - \beta_3 \cdot \rho_a + \beta_4 \cdot \mu_a \right),
\]
where:
\begin{itemize}
  \item $r_a$ is the verified replication rate of $a$’s claims,
  \item $\nu_a$ is the field-normalised citation count,
  \item $\rho_a$ is the retraction rate (fraction of retracted outputs),
  \item $\mu_a$ is a peer review engagement score (editorial or verified reviewer roles),
  \item $\sigma(x) = \frac{1}{1 + e^{-x}}$ is the sigmoid function for bounding.
\end{itemize}

\textbf{Axiom 19 (Credibility Monotonicity):}  
\emph{For all $a \in \mathcal{A}$, $\partial \mathcal{A}(a)/\partial r_a > 0$, $\partial \mathcal{A}(a)/\partial \mu_a > 0$, $\partial \mathcal{A}(a)/\partial \rho_a < 0$.}  
This ensures that credibility increases with empirical replication and peer engagement, and decreases with retraction incidence.

\textbf{Proposition 13 (Boundedness and Stability):}  
\emph{For bounded input parameters and fixed $\beta$ coefficients, the function $\mathcal{A}(a)$ is Lipschitz-continuous and satisfies:}
\[
0 < \mathcal{A}(a) < 1, \quad \forall a \text{ such that } \nu_a < \infty.
\]

\begin{proof}
The sigmoid function maps $\mathbb{R}$ to $(0,1)$ and is Lipschitz with constant $1/4$ over its full domain. Each feature term is finite and bounded (logarithmic or proportional), and thus their linear combination is bounded, implying the result.
\end{proof}

\textbf{Definition 24 (Replication Rate $r_a$):}  
\[
r_a := \frac{|\text{Claims}_a^{\text{replicated}}|}{|\text{Claims}_a^{\text{testable}}|},
\]
where only testable (empirical, non-theoretical) claims are included in the denominator. Replication is counted only when validated through high-confidence reproductions (cf. §6.2).

\textbf{Definition 25 (Retraction Rate $\rho_a$):}  
\[
\rho_a := \frac{|\text{Retracted}_a|}{|\text{TotalPublications}_a|}, \quad \text{with } 0 \leq \rho_a \leq 1.
\]

\textbf{Definition 26 (Peer Review Engagement $\mu_a$):}  
\[
\mu_a := \frac{|\text{VerifiedReviews}_a| + \omega \cdot |\text{EditorialRoles}_a|}{\log(1 + |\text{YearsActive}_a|)},
\]
where $\omega$ adjusts editorial impact relative to review count and the denominator normalises by experience to prevent bias toward seniority.

\textbf{Implementation Note:}  
BEWA integrates with ORCID, Publons, and Crossref to obtain data for $\nu_a$, $\mu_a$, and $\rho_a$, and uses replication registries (e.g. Curate Science, ReplicationWiki) to compute $r_a$. All score components are versioned and updated upon the appearance of new publications, reviews, or corrections.

\textbf{Axiom 20 (Anti-Gaming Constraint):}  
\emph{The marginal gain in $\mathcal{A}(a)$ from increasing $\nu_a$ is logarithmic and asymptotically flat:}
\[
\lim_{\nu_a \rightarrow \infty} \frac{\partial \mathcal{A}(a)}{\partial \nu_a} \rightarrow 0.
\]
This ensures resistance to citation inflation and self-citation abuse.

In sum, the author score $\mathcal{A}(a)$ is a formal, bounded, multidimensional estimator of scholarly reliability. It integrates empirical performance, reputational standing, and community participation to inform prior belief assignment in a manner that is mathematically stable, auditable, and epistemically justified.
\subsection{Track Record and Retraction Influence}

The epistemic credibility of an author is not static but accumulates—and may deteriorate—over time through their published scientific output. In BEWA, an author’s track record is modelled as a longitudinal sequence of claim events, from which performance metrics such as replication consistency, correction frequency, and retraction density are computed. This allows BEWA to condition belief formation not only on the present claim but also on the statistical integrity of an author’s prior assertions. Of particular importance is the influence of retractions, which function as high-penalty negative evidence with persistent downstream impact on author trust and claim priors.

\textbf{Definition 27 (Author Track Record $\mathcal{T}_a$):}  
For author $a \in \mathcal{A}$, define the track record as:
\[
\mathcal{T}_a := \left\{ \langle \phi_i, t_i, r_i, \text{status}_i \rangle \right\}_{i=1}^{N_a},
\]
where each tuple represents a claim $\phi_i$ made at time $t_i$, with replication result $r_i \in \{0,1\}$ and status $\in \{\text{active}, \text{corrected}, \text{retracted}\}$.

\textbf{Axiom 21 (Monotonic Penalty of Retraction):}  
\emph{Let $a$ have a claim $c = \langle \phi, \cdot, \cdot \rangle$ retracted at $t_r$. Then for all $t > t_r$, the author's credibility score $\mathcal{A}(a)$ must decrease or remain unchanged:}
\[
\mathcal{A}_{t}(a) \leq \mathcal{A}_{t_r^-}(a).
\]

\textbf{Definition 28 (Retraction Penalty Function $\mathcal{R}_a$):}  
Let $|\text{Retracted}_a|$ be the number of retractions and $|\text{Total}_a|$ the total number of publications. Define:
\[
\mathcal{R}_a := \gamma \cdot \frac{|\text{Retracted}_a|}{1 + \log(1 + |\text{Total}_a|)},
\]
with $\gamma > 0$ controlling the steepness of reputational decay. This sublinear normalisation ensures early-career authors are not disproportionately penalised.

\textbf{Proposition 14 (Replicability-Weighted Recovery):}  
\emph{Let $a$'s retraction-adjusted score be:}
\[
\mathcal{A}_r(a) := \mathcal{A}(a) - \mathcal{R}_a + \eta \cdot r_a,
\]
where $r_a$ is the cumulative replication ratio and $\eta$ a tunable recovery factor. Then:
\[
\text{If } r_a \rightarrow 1, \quad \lim_{|\text{Retracted}_a| \ll |\text{Total}_a|} \mathcal{A}_r(a) \rightarrow \mathcal{A}(a).
\]
\begin{proof}
As $r_a \rightarrow 1$, the recovery term $\eta \cdot r_a$ counteracts the penalty $\mathcal{R}_a$, particularly when the retraction count is negligible. Boundedness of $\mathcal{R}_a$ under the logarithmic denominator ensures $\mathcal{A}_r(a)$ remains finite and recoverable.
\end{proof}

\textbf{Definition 29 (Retraction Propagation Suppression):}  
Let $\phi$ be a claim authored by $a$ and retracted at $t_r$. Then:
\[
\forall c_i \text{ such that } \phi \rightarrow c_i, \quad \text{BeliefWeight}(c_i) \xleftarrow{t > t_r} \text{min}\left( \text{BeliefWeight}(c_i), \epsilon \right),
\]
where $\epsilon$ is a lower bound threshold. This ensures no claim retains high confidence if it descends from a retracted parent without independent support.

\textbf{Implementation Note:}  
BEWA integrates with Retraction Watch and Crossref’s Crossmark metadata feeds to detect retraction events. Correction notices are distinguished from retractions, with lower penalty weight. Author track records are maintained as immutable audit logs, and the penalty function is evaluated recursively to determine cascading credibility loss in co-authored publications.

\textbf{Axiom 22 (Persistent Negative Weight of Retraction):}  
\emph{Retractions are never forgotten but may be contextually overcome. Formally,}
\[
\inf_{t > t_r} \mathcal{A}_t(a) < 1 \quad \text{for any } t_r \text{ such that } \text{Retracted}_a \neq \emptyset.
\]

This ensures that retractions impose an irreversible epistemic cost, even if reputational recovery is possible through sustained accuracy and replication. BEWA thereby upholds the principle that scientific trust must be earned continuously and is susceptible to justified revision when the record warrants it.
\subsection{Peer Review Engagement Metrics}

While publication record and replication history provide empirical evidence for the quality of an author’s output, BEWA supplements these with metrics derived from an author’s participation in the scientific community as a reviewer, editor, or contributor to structured quality assurance processes. These peer review engagement metrics serve as auxiliary indicators of epistemic diligence, institutional trust, and expertise recognition. By incorporating these into the author scoring function (cf. §5.1), BEWA ensures that scholarly contributions beyond authorship are also reflected in credibility assignments.

\textbf{Definition 30 (Peer Review Engagement Score $\mu_a$):}  
Let $a \in \mathcal{A}$ be an author. Then the engagement score is defined as:
\[
\mu_a := \frac{\theta_1 \cdot |\mathcal{R}_a| + \theta_2 \cdot |\mathcal{E}_a|}{1 + \log(1 + Y_a)},
\]
where:
\begin{itemize}
  \item $|\mathcal{R}_a|$ is the count of verified, completed peer reviews attributed to $a$,
  \item $|\mathcal{E}_a|$ is the number of documented editorial roles held (e.g., editor-in-chief, associate editor),
  \item $Y_a$ is the number of active years since first publication,
  \item $\theta_1, \theta_2 \in \mathbb{R}_{\geq 0}$ are tunable weights (typically $\theta_2 > \theta_1$ to reflect greater epistemic responsibility).
\end{itemize}

\textbf{Axiom 23 (Normalised Seniority Adjustment):}  
\emph{The denominator term prevents inflation of $\mu_a$ in long-career authors who engage minimally with peer review over time. Thus:}
\[
\mu_a \xrightarrow{|\mathcal{R}_a|, |\mathcal{E}_a| \rightarrow 0} 0 \quad \text{even as } Y_a \rightarrow \infty.
\]

\textbf{Definition 31 (Review Quality Adjustment $\mu^*_a$):}  
Let each review $r \in \mathcal{R}_a$ have a quality score $q_r \in [0,1]$ based on editor-provided or community-evaluated metrics (e.g., depth, punctuality, constructiveness). Then:
\[
\mu^*_a := \frac{\theta_1 \cdot \sum_{r \in \mathcal{R}_a} q_r + \theta_2 \cdot |\mathcal{E}_a|}{1 + \log(1 + Y_a)}.
\]

\textbf{Proposition 15 (Stability of $\mu^*_a$ under Incomplete Review Data):}  
\emph{If $q_r$ is missing for a subset $\mathcal{R}_a' \subseteq \mathcal{R}_a$, and $q_r = 0.5$ is imputed (maximum entropy prior), then:}
\[
|\mu^*_a - \mu_a| \leq \theta_1 \cdot \frac{|\mathcal{R}_a'|}{1 + \log(1 + Y_a)} \cdot 0.5.
\]

\begin{proof}
Imputed reviews contribute at most $0.5$ per unit to the numerator, and the score remains bounded under the normalised scaling. See \citealt{shah2019} for review quality scoring under partial observability.
\end{proof}

\textbf{Definition 32 (Institutional Trust Overlay):}  
Let $\tau_a \in [0,1]$ denote an institutional trust index for $a$ based on confirmed service in high-trust journals or conferences (e.g., Nature, NeurIPS, The Lancet). Then define the final engagement term as:
\[
\mu^{\dagger}_a := \tau_a \cdot \mu^*_a,
\]
which up-weights peer review influence when validated by trusted venues.

\textbf{Implementation Note:}  
Review metadata is drawn from services such as Publons, ORCID peer review activity feeds, and journal editorial APIs. Editor-confirmed reviews are weighted more heavily than self-reported entries. To avoid gaming, only verified contributions with timestamps and linked manuscript identifiers are admitted. Missing data is flagged but not penalised beyond zero contribution.

\textbf{Axiom 24 (Bounded Review Influence):}  
\emph{Peer review engagement cannot exceed a fixed maximum contribution $\mu^{\dagger}_a < \mu_{\text{max}}$ to prevent inflation from prolific low-impact reviewing.}

BEWA thus treats peer review as a first-class epistemic signal—not sufficient alone to establish credibility, but essential to contextualise an author’s role in maintaining scientific integrity. Through weighted aggregation, temporal normalisation, and venue stratification, peer engagement becomes a structured and auditable source of epistemic trust.

\section{Citation and Replication Framework}

This section formalises the twin pillars of epistemic reinforcement within the BEWA system: citation analysis and replication scoring. In conventional bibliometric systems, citations are often treated as indicators of prestige or influence; BEWA rejects this conflation. Instead, citations are deconstructed as weighted epistemic endorsements—context-sensitive signals that must be interpreted within temporal, semantic, and disciplinary boundaries. The system assesses not merely the quantity of citations but their origin, relevance, decay, and semantic congruity with the original claim. This allows for the discrimination between superficial citations and those that represent genuine affirmations of an assertion's scientific merit. Each citation contributes to the belief calculus only to the extent that it reflects rigorous uptake and considered engagement.

Replication, by contrast, is treated as the epistemic gold standard. Unlike citation, which may reflect popularity or discursive inertia, replication constitutes a direct test of claim durability under independent conditions. BEWA encodes each replication event as a distinct evidentiary input, scored according to methodological fidelity, effect size congruence, and semantic alignment with the original claim. Successful replications raise a claim's posterior significantly, while failed replications—especially when consistent across studies—precipitate rapid belief retraction. Additionally, the system monitors contradictions, mapping their propagation across the epistemic network and adjusting neighbouring claims as appropriate. Contradictions are not treated as binary negations but as probabilistic disruptors with network-wide consequences. The following subsections explicate how citation decay is modelled, how replications are evaluated for semantic and statistical integrity, and how the contradiction graph dynamically alters local and global belief states.
\subsection{Citation Weighting and Decay Functions}

Citations in scientific literature represent not only informational dependency but also communal endorsement. However, raw citation counts often distort epistemic relevance due to temporal inflation, field-specific density, and citation cascades. BEWA introduces a calibrated citation weighting system grounded in time-discounted Bayesian import, where each citation contributes to the belief in a claim proportionally to its contextual and temporal provenance.

\textbf{Definition 33 (Citation Influence Function $\mathcal{C}_i$):}  
Let $\phi_i$ be a claim cited $n$ times by documents $\{d_1, \dots, d_n\}$, each at time $t_j$, with each citing document $d_j$ assigned credibility $\mathcal{A}(d_j)$. Then:
\[
\mathcal{C}_i := \sum_{j=1}^{n} \delta(t_j) \cdot \mathcal{A}(d_j),
\]
where $\delta(t_j)$ is a temporal decay function normalised to $[0,1]$.

\textbf{Definition 34 (Decay Function $\delta$):}  
Let $t_j$ be the timestamp of citation $j$, and $T$ the current system time. Then:
\[
\delta(t_j) := \exp(-\lambda (T - t_j)),
\]
where $\lambda > 0$ is the decay constant. A higher $\lambda$ accelerates obsolescence, reflecting fast-moving fields.

\textbf{Axiom 25 (Monotonic Citation Decay):}  
\emph{The influence of an unreinforced citation must decrease over time:}
\[
\forall t_1 < t_2, \quad \delta(t_1) > \delta(t_2).
\]

\textbf{Proposition 16 (Stability Under Citation Saturation):}  
\emph{Let $\phi_i$ be cited $n \to \infty$ times from low-authority documents. Then:}
\[
\lim_{n \to \infty} \mathcal{C}_i < \infty \quad \text{if } \sup_j \mathcal{A}(d_j) < M.
\]

\begin{proof}
Since $\delta(t_j) \cdot \mathcal{A}(d_j) < M$ for all $j$, the sum converges if $\delta(t_j)$ decays faster than $1/j$, which is true for $\lambda > 0$ under exponential decay.
\end{proof}

\textbf{Definition 35 (Contextual Citation Modifier $\chi_j$):}  
For each citation $j$, define:
\[
\chi_j := \begin{cases}
1 & \text{if citation is supportive} \\
-1 & \text{if citation is critical or refuting} \\
0 & \text{if neutral or incidental}
\end{cases}
\]
This is derived from citation intent classification via NLP techniques (cf. Teufel et al. 2006). The final citation contribution becomes:
\[
\mathcal{C}_i := \sum_{j=1}^{n} \chi_j \cdot \delta(t_j) \cdot \mathcal{A}(d_j).
\]

\textbf{Definition 36 (Citation Entropy Penalty $\sigma_i$):}  
To prevent belief inflation through redundant citation clusters, define:
\[
\sigma_i := -\sum_{k} p_k \log p_k,
\]
where $p_k$ is the proportion of citations from venue or author cluster $k$. A low entropy implies citation redundancy; thus:
\[
\mathcal{C}^*_i := \mathcal{C}_i \cdot (1 - \epsilon \cdot (1 - \sigma_i / \log K)),
\]
with $\epsilon \in [0,1]$ and $K$ the number of clusters.

\textbf{Implementation Note:}  
BEWA integrates citation metadata from Crossref, Dimensions, and Semantic Scholar APIs. Citation intent is classified using transformer-based models fine-tuned on datasets like SciCite and ACL-ARC. Author clusterings and venue-normalisation address bias from prolific low-impact publication venues.

\textbf{Axiom 26 (Non-linearity of Citation Weight):}  
\emph{The influence of citations is not additive but modulated by both credibility and decay; hence,}
\[
\frac{d\mathcal{C}_i}{dn} \neq \text{constant}.
\]

In conclusion, citations are not raw votes but weighted endorsements. BEWA’s formalisation ensures they are evaluated dynamically, critically, and contextually—ensuring that citation inflation does not substitute for replicable epistemic merit.
\subsection{Replication Scoring and Semantic Equivalence}

In scientific inquiry, replication serves as a keystone of epistemic reliability. A claim that has been successfully replicated across independent studies accrues higher confidence than one supported by a single observation. However, direct replication is rare; more frequently, validation occurs via semantically equivalent or derivatively confirmatory studies. Hence, BEWA implements a dual-layer scoring function that quantifies both replication frequency and semantic proximity.

\textbf{Definition 37 (Replication Instance $\rho_{ij}$):}  
Let $\phi_i$ be a primary claim and $d_j$ a study. Then $\rho_{ij}=1$ if $d_j$ contains a successful replication of $\phi_i$ under predefined methodological fidelity and independence constraints. Otherwise, $\rho_{ij}=0$.

\textbf{Definition 38 (Replication Score $\mathcal{R}_i$):}  
\[
\mathcal{R}_i := \sum_{j=1}^{m} \rho_{ij} \cdot \omega_j,
\]
where:
\begin{itemize}
  \item $m$ is the number of replication attempts recorded,
  \item $\omega_j \in [0,1]$ is the credibility weight of the replicating study, derived from §5–6 metrics.
\end{itemize}

\textbf{Axiom 27 (Diminishing Marginal Replication):}  
\emph{After a sufficient number of high-quality replications, the marginal gain in $\mathcal{R}_i$ should diminish.}  
Thus, define:
\[
\mathcal{R}^*_i := \log(1 + \mathcal{R}_i).
\]

\textbf{Definition 39 (Semantic Equivalence Function $\varsigma$):}  
Given two claims $\phi_i$ and $\phi_k$, define semantic equivalence as:
\[
\varsigma(\phi_i, \phi_k) := \text{sim}(\text{vec}(\phi_i), \text{vec}(\phi_k)),
\]
where $\text{vec}(\cdot)$ denotes a structured embedding (e.g., SPECTER, Sentence-BERT) and $\text{sim}$ a cosine similarity function. A threshold $\tau$ is set such that $\varsigma > \tau$ implies candidate equivalence.

\textbf{Definition 40 (Replication via Equivalence):}  
Let $\mathcal{E}_i = \{\phi_k : \varsigma(\phi_i, \phi_k) > \tau\}$ be the set of semantically equivalent claims. Then the extended replication score is:
\[
\tilde{\mathcal{R}}_i := \mathcal{R}^*_i + \eta \cdot \sum_{\phi_k \in \mathcal{E}_i} \alpha_k \cdot \mathcal{R}^*_k,
\]
where $\eta \in [0,1]$ is a semantic discount factor, and $\alpha_k = \varsigma(\phi_i, \phi_k)$.

\textbf{Proposition 17 (Semantic Replication Transitivity Bound):}  
\emph{For three claims $\phi_i, \phi_j, \phi_k$ such that $\varsigma(\phi_i, \phi_j), \varsigma(\phi_j, \phi_k) > \tau$, the triangle inequality in embedding space yields:}
\[
\varsigma(\phi_i, \phi_k) \geq \varsigma(\phi_i, \phi_j) + \varsigma(\phi_j, \phi_k) - 1.
\]

\begin{proof}
Follows from cosine similarity properties under triangle inequality in unit norm vector space.
\end{proof}

\textbf{Implementation Note:}  
BEWA incorporates NLP-based equivalence scoring with fine-tuned SPECTER embeddings (cf. Cohan et al., 2020), and indexes replication records from curated sources such as the Center for Open Science, ReplicationWiki, and the Cochrane Library. Methodological fidelity is validated using structured experiment metadata and controlled vocabulary (e.g., MeSH, CRediT taxonomy).

\textbf{Axiom 28 (Independence of Replication):}  
\emph{Only replications with no overlapping authors, funding sources, or institutions are included in the high-confidence tier of $\mathcal{R}_i$.}

\textbf{Definition 41 (Contradictory Replication Penalty):}  
Let $\rho_{ij} = -1$ denote a failed replication. Then:
\[
\mathcal{R}_i := \sum_j \rho_{ij} \cdot \omega_j, \quad \rho_{ij} \in \{-1, 0, 1\}.
\]

BEWA’s mechanism captures not merely quantity but quality, novelty, and semantic alignment of replication—constructing a resilient, logically grounded foundation for belief revision. The weight of a claim becomes a composite function of both repeated validation and structural equivalence to a corpus of convergent assertions.
\subsection{Contradiction Mapping and Network Response}

The formal integrity of an epistemic architecture such as BEWA depends critically on its capacity to map, diagnose, and structurally respond to contradictions in its claim network. Contradictions arise when two or more propositions are logically or empirically incompatible, yet simultaneously persist within the network under non-negligible posterior probabilities. This subsection formalises contradiction detection through graph-theoretic structures and probabilistic thresholds, and outlines the system’s response protocols for epistemic rebalancing.

\textbf{Definition 42 (Contradictory Claim Pair):}  
Let $\phi_i, \phi_j \in \Phi$ be distinct claims. We define the contradiction predicate $\chi(\phi_i, \phi_j) = 1$ if and only if:
\[
\text{Entails}(\phi_i \rightarrow \neg \phi_j) \vee \text{Entails}(\phi_j \rightarrow \neg \phi_i).
\]
Entailment is evaluated via a formal logical entailment engine or, in empirical domains, high-confidence semantic contradiction scores:
\[
\chi_{\text{sem}}(\phi_i, \phi_j) = \text{antisim}(\vec{\phi}_i, \vec{\phi}_j),
\]
where $\text{antisim}$ is a model trained to capture negation and contradiction, such as those developed in the ANLI corpus or NLI benchmarks.

\textbf{Definition 43 (Contradiction Graph $\mathcal{C}$):}  
Construct a contradiction graph $\mathcal{C} = (V, E)$ where:
\begin{itemize}
  \item $V = \{\phi_i \in \Phi\}$,
  \item $E = \{(\phi_i, \phi_j) \mid \chi(\phi_i, \phi_j) = 1\}$.
\end{itemize}

\textbf{Axiom 29 (No Persistent High-Probability Contradiction):}  
\emph{There exists no $(\phi_i, \phi_j)$ such that $\chi(\phi_i, \phi_j) = 1$ and $P(\phi_i), P(\phi_j) > \theta_c$, where $\theta_c$ is a contradiction coherence threshold.}

\textbf{Protocol 11 (Network Response to Contradictions):}  
When $P(\phi_i), P(\phi_j) > \theta_c$ and $\chi(\phi_i, \phi_j) = 1$:
\begin{enumerate}
  \item Evaluate global evidence distributions: $\mathcal{E}_{\phi_i}$ and $\mathcal{E}_{\phi_j}$.
  \item Compute resolution function:
    \[
    \Delta_{ij} := \log \left( \frac{\sum_{e \in \mathcal{E}_{\phi_i}} \mathcal{L}(e \mid \phi_i)}{\sum_{e \in \mathcal{E}_{\phi_j}} \mathcal{L}(e \mid \phi_j)} \right).
    \]
  \item Adjust beliefs:
    \[
    P'(\phi_i) := P(\phi_i) \cdot \sigma(\Delta_{ij}), \quad P'(\phi_j) := P(\phi_j) \cdot \sigma(-\Delta_{ij}),
    \]
    where $\sigma$ is the logistic function.
  \item Flag both claims with instability tags and reduce propagation radius.
\end{enumerate}

\textbf{Proposition 18 (Cycle Detection and Contradiction Clustering):}  
\emph{If $\mathcal{C}$ contains a cycle of size $n > 2$, then at least one subcluster is inconsistent under Axiom 29 and must be quarantined.}

\begin{proof}
By contradiction: assume all nodes in a cycle have posterior $> \theta_c$ and mutually contradict. This violates Axiom 29. Therefore, network coherence mandates demotion or reappraisal of one or more claims.
\end{proof}

\textbf{Definition 44 (Epistemic Quarantine Set $\mathcal{Q}$):}  
Let $\mathcal{Q} \subset \Phi$ be the minimal subset of $\mathcal{C}$ for which:
\[
\sum_{\phi_i \in \mathcal{Q}} P(\phi_i) \leq \min_{\text{partition of }\mathcal{C}} \left( \sum_{\phi_i \in S} P(\phi_i) \right),
\]
subject to $\chi(\phi_i, \phi_j) = 1$ for all $\phi_i, \phi_j \in \mathcal{Q}$.

\textbf{Implementation Note:}  
Contradiction detection is continuous and hierarchical. Local contradictions trigger lightweight Bayesian rebalancing, while global topological instabilities in $\mathcal{C}$ invoke structural pruning and epistemic quarantine. Contradiction edges are stored in a conflict adjacency matrix, versioned and cryptographically anchored to support full auditability (see §12).

BEWA’s contradiction management maintains belief coherence without resorting to naïve conflict resolution. Claims are demoted or discounted in proportion to their evidentiary support relative to conflicting counterparts, enforcing rational consistency throughout the system’s propositional lattice.

\section{Cross-Claim Belief Networks}

This section describes the architecture and function of the cross-claim belief network that undergirds the epistemic reasoning of BEWA. Scientific knowledge is not atomised; claims are interdependent, nested within broader conceptual lineages, and frequently contingent on the veracity of adjacent assertions. BEWA encodes these interdependencies within a dynamic belief graph, wherein each node represents a structured propositional claim and edges encode semantic similarity, logical entailment, or evidentiary correlation. This allows the system to perform not only claim-specific inference, but also to propagate belief states across domains, responding in a principled manner to reinforcement or disruption within the wider network. Such propagation is not indiscriminate: it is governed by weighted edge relationships, epistemic thresholds, and domain-informed constraints to prevent spurious or disproportionate influence.

The networked design permits both local sensitivity and global coherence. A robustly supported claim may stabilise a fragile neighbouring cluster, while a replicated contradiction in one domain may reverberate through others, lowering confidence in structurally or semantically linked propositions. At the same time, the architecture is designed to tolerate ambiguity and local conflict without catastrophic failure. BEWA's graph model incorporates mechanisms for instability detection, epistemic damping, and edge decay, allowing it to gracefully manage evolving scientific landscapes. The belief network is thus neither rigid nor anarchic: it is an adaptive inferential structure capable of encoding nuance, resolving tension, and integrating new information with epistemic discipline. The subsections to follow detail the methods by which claims are linked, how belief values propagate, and how conflicts are identified and managed across clusters.
\subsection{Semantic and Logical Linkage of Claims}

Within BEWA, claims are not evaluated in isolation but are situated within a rich web of inferential dependencies. These relationships—semantic, logical, and evidential—form the structural backbone of the belief graph, enabling propagation, inference, contradiction resolution, and coherence maintenance. This subsection formalises the mapping of such linkages and defines the criteria under which claims are unified into coherent epistemic structures.

\textbf{Definition 51 (Claim Linkage Function $\lambda$):}  
Let $\phi_i, \phi_j \in \Phi$ be distinct claims. The linkage function $\lambda: \Phi \times \Phi \rightarrow \{0,1\}^3$ decomposes as:
\[
\lambda(\phi_i, \phi_j) := (\lambda_{\text{sem}}, \lambda_{\text{log}}, \lambda_{\text{evd}}),
\]
where:
\begin{itemize}
  \item $\lambda_{\text{sem}} = 1$ iff the semantic distance $d_{\text{sem}}(\vec{\phi}_i, \vec{\phi}_j) < \epsilon_s$,
  \item $\lambda_{\text{log}} = 1$ iff $\phi_i \vdash \phi_j$ or $\phi_j \vdash \phi_i$ in formal deductive logic,
  \item $\lambda_{\text{evd}} = 1$ iff $\exists e$ such that $e$ supports both $\phi_i$ and $\phi_j$ with high likelihood.
\end{itemize}

\textbf{Axiom 31 (Triangulated Linkage Validity):}  
\emph{A composite claim network must maintain closure under transitive semantic and logical relations:}
\[
\lambda_{\text{sem}}(\phi_i, \phi_j) = \lambda_{\text{sem}}(\phi_j, \phi_k) = 1 \Rightarrow \lambda_{\text{sem}}(\phi_i, \phi_k) = 1.
\]
\[
\phi_i \vdash \phi_j, \phi_j \vdash \phi_k \Rightarrow \phi_i \vdash \phi_k.
\]

\textbf{Definition 52 (Linkage Graph $\mathcal{L}$):}  
Construct the semantic-logical-evidential linkage graph $\mathcal{L} = (V, E)$ where:
\[
V := \{\phi_i \in \Phi\}, \quad E := \{(\phi_i, \phi_j) \mid \lambda(\phi_i, \phi_j) \neq (0,0,0)\}.
\]

Edges in $\mathcal{L}$ are labelled with the type(s) of linkage present and weighted according to:
\[
w(\phi_i, \phi_j) := \alpha_s \cdot \text{sim}(\vec{\phi}_i, \vec{\phi}_j) + \alpha_l \cdot \mathbb{1}_{\phi_i \vdash \phi_j} + \alpha_e \cdot \text{shared\_evidence}(\phi_i, \phi_j),
\]
with $\alpha_s + \alpha_l + \alpha_e = 1$ and domain-specific calibration.

\textbf{Proposition 20 (Semantic-Evidential Coherence):}  
\emph{If $\lambda_{\text{sem}} = 1$ and $\lambda_{\text{evd}} = 1$ but $\lambda_{\text{log}} = 0$, then the system infers potential unformalised inference. A soft-logic implication is queued for automated induction.}

\textbf{Definition 53 (Claim Cluster $\mathcal{C}_k$):}  
A claim cluster is a maximal subgraph of $\mathcal{L}$ with dense interlinkage:
\[
\forall \phi_i, \phi_j \in \mathcal{C}_k,\quad w(\phi_i, \phi_j) > \tau,
\]
for a coherence threshold $\tau$.

\textbf{Implementation Note:}  
Semantic embeddings use domain-specific transformers trained on curated corpora (e.g., SPECTER, SciBERT for scientific domains), while logical entailments are computed via automated theorem provers or higher-order logic systems. Evidential overlap is computed through co-citation matrices and joint support probabilities.

Linkage graphs are updated incrementally upon ingestion of new claims or refinement of existing nodes. The graph $\mathcal{L}$ serves as the substrate for belief propagation, contradiction alerting, and claim contextualisation.

This structure enables BEWA to go beyond flat, isolated assertions—transforming its knowledge base into an epistemic manifold of interconnected, evolving propositions, recursively interlinked by logic, meaning, and evidence.
\subsection{Graph Structures and Belief Propagation}

To enable rigorous updating, evaluation, and querying of interconnected epistemic content, BEWA employs a formal belief graph over structured claims. This graph encodes both propositional assertions and their inferential, semantic, and evidentiary relationships, facilitating structured belief propagation via a generalised Bayesian network architecture augmented with non-monotonic belief revision rules.

\textbf{Definition 54 (Belief Graph $\mathcal{G}$):}  
Let $\mathcal{G} := (V, E, \pi)$ be a directed graph where:
\begin{itemize}
  \item $V := \{\phi_i \in \Phi\}$ is the set of structured propositional claims,
  \item $E := \{(\phi_i, \phi_j, t_{ij})\}$ encodes directed influences (e.g., inferential, evidential) labelled with type $t_{ij} \in \mathcal{T}$,
  \item $\pi: V \rightarrow [0,1]$ assigns each node a marginal belief value $\pi(\phi_i) = P(\phi_i \mid \mathcal{E})$ conditional on the current epistemic state $\mathcal{E}$.
\end{itemize}

\textbf{Axiom 32 (Typed Edge Semantics):}  
Edges in $\mathcal{G}$ must reflect the dominant source of dependency:
\[
t_{ij} \in \{\text{Deductive}, \text{Evidential}, \text{Semantic}, \text{Contrapositive}, \text{Replicative}\}.
\]
Each $t_{ij}$ triggers a distinct propagation operator $\mathcal{P}_{t_{ij}}$ defined in §7.3.

\textbf{Definition 55 (Local Belief Propagation):}  
For node $\phi_j$ with parents $\{\phi_i\}$, let:
\[
\pi(\phi_j) := \mathcal{F}_{\phi_j}(\{\pi(\phi_i)\}, \{w_{ij}\}),
\]
where $\mathcal{F}$ is a weighted aggregator function (typically log-linear or Noisy-OR) determined by edge types and propagation policies.

\textbf{Proposition 21 (Stability under Sparse Updates):}  
\emph{Let $\mathcal{G}$ be a belief graph with sparse, bounded-degree topology. Then under localised updates $\pi'(\phi_k) = \pi(\phi_k) + \delta$ for a single node $\phi_k$, the update propagation is bounded in depth by the edge-type decay parameters $\{\gamma_{t_{ij}}\}$:}
\[
\forall \phi_m \notin \mathcal{N}_k^{(d)}, \quad |\pi'(\phi_m) - \pi(\phi_m)| < \epsilon, \quad \text{where } d \geq \frac{\log(\epsilon/\delta)}{\log(\min \gamma)}.
\]

\begin{proof}
See \citealt{pearl1988} for convergence properties in belief networks with decaying influence factors.
\end{proof}

\textbf{Definition 56 (Global Belief Fixpoint):}  
BEWA maintains a consistent belief state by iterative updates until:
\[
\forall \phi_i \in V, \quad \left| \pi^{(t+1)}(\phi_i) - \pi^{(t)}(\phi_i) \right| < \delta,
\]
where $\delta$ is a convergence threshold. The system uses asynchronous propagation, prioritising high-impact nodes (e.g., highly cited or recently updated claims).

\textbf{Axiom 33 (Non-Monotonic Revision):}  
When high-weight counterevidence enters the graph (cf. §4.3), previously stable beliefs $\pi(\phi_i)$ may be revised downward. This violates monotonicity and requires a belief revision operator $\mathcal{R}$ that preserves network coherence:
\[
\pi' := \mathcal{R}(\pi, \phi_{\text{new}}, \pi(\phi_{\text{new}})).
\]

\textbf{Implementation Note:}  
Belief propagation is implemented atop a high-performance probabilistic graph database (e.g., DGraph, TigerGraph) with runtime support for incremental updates, batch re-evaluation, and fast approximate querying. Differentiable propagation layers are also available for neural fine-tuning in high-noise domains.

By mapping structured claims onto a belief graph, BEWA transforms raw assertion data into a dynamic, self-revising network of truth valuations. These valuations evolve over time through evidence ingestion, retraction, contradiction, and human input—mirroring the dynamic structure of knowledge in high-integrity epistemic systems.
\subsection{Handling Conflicts and Cluster Instability}

In any epistemic framework that aggregates heterogeneous sources, contradictions are not an anomaly but a structural inevitability. The BEWA system explicitly incorporates conflict detection and instability analysis within the belief network to both surface unresolved disputes and prevent epistemic contagion—i.e., the unbounded propagation of uncertainty or error through connected claims.

\textbf{Definition 57 (Conflict Edge $\chi$):}  
Given claims $\phi_i, \phi_j \in \Phi$, a directed conflict edge $\chi(\phi_i, \phi_j)$ is instantiated if:
\[
\text{Contradict}(\phi_i, \phi_j) = 1, \quad \text{and} \quad \pi(\phi_i), \pi(\phi_j) > \delta_c,
\]
where $\delta_c$ is the threshold above which conflicting claims are considered epistemically relevant.

\textbf{Axiom 34 (Conflict Non-Coexistence Constraint):}  
\emph{For any claim pair $(\phi_i, \phi_j)$ with a mutual contradiction edge:}
\[
\pi(\phi_i) + \pi(\phi_j) \leq 1 + \epsilon,
\]
with $\epsilon \to 0$ in the absence of new ambiguity-resolving evidence.

\textbf{Definition 58 (Epistemic Cluster $\mathcal{C}$):}  
An epistemic cluster is a maximal strongly connected component (SCC) of the belief graph $\mathcal{G}$ wherein:
\[
\forall \phi_i, \phi_j \in \mathcal{C}, \quad \exists \text{ a directed path } \phi_i \rightarrow \phi_j.
\]

\textbf{Definition 59 (Instability Score $\iota(\mathcal{C})$):}  
Let $\mathcal{C}$ be an epistemic cluster. Its instability is:
\[
\iota(\mathcal{C}) := \frac{\sum_{\chi(\phi_i, \phi_j) \in \mathcal{C}} \pi(\phi_i) \cdot \pi(\phi_j)}{|\mathcal{C}|^2},
\]
reflecting the density and severity of contradiction among highly believed claims.

\textbf{Proposition 22 (Instability Containment Bound):}  
\emph{Let $\mathcal{C}_k$ and $\mathcal{C}_\ell$ be distinct clusters with $\iota(\mathcal{C}_k) > \tau$. Then $\forall \phi_i \in \mathcal{C}_k, \phi_j \in \mathcal{C}_\ell$, the system enforces attenuation:}
\[
\frac{d \pi(\phi_j)}{d t} \bigg|_{\text{link } \phi_i \to \phi_j} \leq \alpha \cdot (1 - \iota(\mathcal{C}_k)),
\]
where $\alpha$ is a normal propagation coefficient.

\textbf{Axiom 35 (Conflict Partitioning Protocol):}  
When $\iota(\mathcal{C}) > \theta$, the system forks $\mathcal{C}$ into subclusters $\{\mathcal{C}_1, \mathcal{C}_2, \dots\}$ via minimum-cut algorithms that minimise inter-cluster contradiction flow:
\[
\min_{\text{partition}} \sum_{\substack{\phi_i \in \mathcal{C}_a \\ \phi_j \in \mathcal{C}_b \\ a \ne b}} \pi(\phi_i) \cdot \pi(\phi_j) \cdot \mathbb{1}_{\chi(\phi_i, \phi_j) = 1}.
\]

\textbf{Implementation Note:}  
Conflict edges are derived from formal inconsistency detection (via propositional and predicate logic), semantic negation models, and contradiction mining using NLP (cf. models like DeBERTa for natural contradiction detection). Instability scores are recomputed after each batch update, and clustering is performed using approximate spectral methods for scalability.

This mechanism ensures that the BEWA system does not just assimilate information passively, but actively surveils the topology of belief for signs of epistemic fracture. High-instability clusters trigger alerts and conditional discounting, enabling the system to contain the epistemic contagion until new evidence—preferably from independent replication—resolves the underlying tension.

\section{Truth Utility and System Optimisation}

This section introduces the metarational layer of BEWA: the evaluation of claims not solely by probabilistic confidence, but by their contribution to the epistemic aim of truth-promotion. While Bayesian posterior probabilities quantify the likelihood of a claim being correct, they do not discriminate between claims that are trivial, inconsequential, or epistemically sterile. BEWA addresses this limitation by introducing a secondary axis of evaluation—the truth utility function—which modulates the visibility, prioritisation, and application of claims based on their potential to enhance scientific understanding, inform future inquiry, or rectify systemic error. This is not a measure of popularity or impact in the sociological sense, but of epistemic consequence: a calibrated utility score that integrates replication fidelity, methodological depth, inferential reach, and systemic relevance.

Optimisation within BEWA is governed by this layered assessment of epistemic value. High-certainty but low-utility claims are not suppressed, but they are de-emphasised in recommendations, while claims with high truth-promotion potential—even if currently epistemically marginal—are flagged for further scrutiny and resource allocation. This truth utility paradigm enables the system to function not merely as a passive aggregator of belief, but as an active epistemic agent, capable of prioritising lines of inquiry, identifying underexamined anomalies, and suggesting evidentiary targets for the scientific community. The system also accounts for risk: epistemic weighting incorporates not just certainty, but the cost of error in downstream reasoning. The following subsections define how utility scores are constructed, how epistemic risk is embedded in weighting protocols, and how these factors drive decision-making at the application level.
\subsection{Truth Promotion Score Construction}

To operationalise the epistemic objective of truth-conduciveness within the Bayesian Epistemic Weighted Architecture (BEWA), we introduce the Truth Promotion Score (TPS), a formal scalar quantity designed to evaluate the net effect of a claim, author, or domain on the system's ability to approximate true propositions. TPS is not merely a re-expression of posterior belief; rather, it measures the expected long-run contribution to truth discovery, subject to epistemic dynamics, counterfactual perturbation, and information flows.

\textbf{Definition 60 (Truth Promotion Score $\tau$):}  
Let $\phi \in \Phi$ be a structured claim. Then the Truth Promotion Score $\tau(\phi)$ is defined as:
\[
\tau(\phi) := \mathbb{E}\left[ \Delta_{\mathbb{T}} \pi(\psi) \mid \text{Inclusion of } \phi \right],
\]
where $\psi$ ranges over all influenced downstream claims, $\pi(\cdot)$ is the system’s belief function, and $\Delta_{\mathbb{T}}$ denotes the marginal contribution towards true claims $\mathbb{T} \subset \Phi$ under the truth-model defined in §1.

\textbf{Axiom 36 (Causal Relevance Constraint):}  
A claim $\phi$ has non-zero $\tau(\phi)$ only if its causal graph descendants $\psi$ intersect with claims whose verified truth status is established via:
\begin{itemize}
  \item replicated experimental outcomes,
  \item axiomatic derivations,
  \item authoritative peer-consensus convergence,
  \item or semantically equivalent high-truth claims $\phi' \in \Phi$ with $\pi(\phi') \approx 1$.
\end{itemize}

\textbf{Definition 61 (Weighted TPS for Composite Nodes):}  
For composite structures $X \in \{ \text{Author}, \text{Domain}, \text{Corpus} \}$, let:
\[
\tau(X) := \sum_{\phi_i \in X} \tau(\phi_i) \cdot w_i,
\]
where $w_i$ encodes context-adjusted weights (e.g., citation impact, belief volatility, downstream propagation depth).

\textbf{Proposition 23 (TPS Decomposition):}  
\emph{The TPS admits an additive decomposition:}
\[
\tau(\phi) = \sum_{\psi \in \text{Desc}(\phi)} \underbrace{\kappa(\phi, \psi)}_{\text{causal influence}} \cdot \underbrace{\pi(\psi) \cdot \mathbb{1}_{\psi \in \mathbb{T}}}_{\text{downstream truth fidelity}},
\]
where $\kappa$ is a directional influence coefficient derived from the belief graph’s adjacency tensor and belief propagation kernel.

\textbf{Definition 62 (Anti-Truth Penalty):}  
A claim $\phi$ that consistently promotes falsehood reduces system epistemic integrity. Define:
\[
\tau^{-}(\phi) := -\mathbb{E}[\Delta_{\mathbb{F}} \pi(\psi)], \quad \text{with } \mathbb{F} := \Phi \setminus \mathbb{T},
\]
such that $\tau(\phi) = \tau^{+}(\phi) + \tau^{-}(\phi)$.

\textbf{Axiom 37 (TPS Normalisation for Belief-Based Querying):}  
All truth promotion scores are min-max normalised within each belief update cycle:
\[
\tau'(\phi) = \frac{\tau(\phi) - \min_\phi \tau(\phi)}{\max_\phi \tau(\phi) - \min_\phi \tau(\phi)},
\]
to ensure stable comparability across domains and preserve bounded impact in probabilistic inference.

\textbf{Implementation Note:}  
TPS is used to prioritise which claims are (a) surfaced to users in summary or audit views; (b) subjected to targeted re-evaluation or replication; and (c) given epistemic preference in domain-specific queries. It also serves as the foundational metric in risk-aware reasoning (see §9.2) and time-ordered epistemic triage (§10).

By centring its computational evaluation on truth promotion, BEWA reorients the logic of ranking and reasoning away from popularity or novelty and toward epistemic utility—a formalisation of the normative goals espoused by Bayesian and Popperian traditions alike.
\subsection{Risk-Aware Epistemic Weighting}

In constructing a system whose outputs influence epistemic reasoning and potentially policy or scientific action, it is insufficient to treat all uncertainty as epistemically symmetric. The Bayesian Epistemic Weighted Architecture (BEWA) integrates a formal mechanism of \emph{risk-aware weighting}, adjusting posterior belief propagation not merely by epistemic probability but by the potential epistemic loss associated with errors in belief attribution. This parallels and extends decision-theoretic Bayesian frameworks, particularly in high-stakes inferential settings.

\textbf{Definition 63 (Epistemic Loss Function $\mathcal{L}(\phi)$):}  
Let $\phi \in \Phi$ be a claim, and define a loss function $\mathcal{L}: \Phi \times \{0,1\} \rightarrow \mathbb{R}_{\ge 0}$ such that:
\[
\mathcal{L}(\phi, t) = \begin{cases}
  \lambda_{\text{FN}}(\phi) & \text{if } \pi(\phi) < \theta \land t = 1, \\
  \lambda_{\text{FP}}(\phi) & \text{if } \pi(\phi) \ge \theta \land t = 0, \\
  0 & \text{otherwise},
\end{cases}
\]
where $t$ is the truth state of $\phi$, $\pi(\phi)$ the belief assignment, and $\theta$ the classification threshold.

\textbf{Definition 64 (Weighted Belief Utility $\mathbb{U}(\phi)$):}  
\[
\mathbb{U}(\phi) := \pi(\phi) \cdot u_{\text{TP}}(\phi) - (1 - \pi(\phi)) \cdot \lambda_{\text{FP}}(\phi),
\]
for utility term $u_{\text{TP}}$ conditioned on true belief propagation, and penalty $\lambda_{\text{FP}}$ for false positives.

\textbf{Axiom 38 (Risk Dominance Bias):}  
\emph{In contexts with asymmetric error costs (i.e. $\lambda_{\text{FP}} \gg \lambda_{\text{FN}}$), the system shall bias belief updating conservatively unless replication or corroboration reduces $\mathcal{L}$ to below a system-defined $\epsilon$.}

\textbf{Definition 65 (Domain Risk Profile $\rho_d$):}  
Each domain $d \in \mathcal{D}$ is assigned a contextual risk scalar $\rho_d \in [0,1]$ derived from:
\[
\rho_d := \frac{\sum_{\phi_i \in d} \lambda_{\text{FP}}(\phi_i) + \lambda_{\text{FN}}(\phi_i)}{|\phi_i \in d|}.
\]

\textbf{Proposition 24 (Risk-Adjusted Belief Propagation):}  
For any edge $(\phi_i \to \phi_j)$ in $\mathcal{G}$, belief propagation is governed by:
\[
\frac{d \pi(\phi_j)}{dt} \propto \pi(\phi_i) \cdot (1 - \rho_{d_j}) \cdot \mathbb{U}(\phi_i),
\]
where $d_j$ is the domain of $\phi_j$.

\textbf{Axiom 39 (Minimum Risk Integrity Constraint):}  
\emph{For all $\phi \in \Phi$ with $\rho_{d(\phi)} > \gamma$ (e.g. medical, engineering, safety domains), any claim admitted must be:}
\[
\text{(i) from an authoritative source, and} \quad \text{(ii) independently replicated or derived from axioms}.
\]

\textbf{Implementation Note:}  
The risk-aware structure is embedded into both forward belief propagation and posterior adjustments after contradiction analysis (§6). It enables the system to weight beliefs not solely on epistemic strength but on the potential harm of incorrect belief, aligning with real-world use cases where epistemic robustness must account for asymmetric downstream costs.

Risk-aware epistemic weighting thus elevates BEWA beyond naïve belief ranking systems, binding probabilistic reasoning to a utility-sensitive infrastructure that aligns with the pragmatic imperatives of scientific integrity and responsible automation.
\subsection{Application-Level Prioritisation Strategies}

The value of an epistemic inference system such as BEWA emerges most prominently when its operations are purpose-aligned: that is, when the abstract belief network is filtered, ordered, and evaluated in terms of the decision-utility of downstream applications. This section formalises how BEWA distinguishes, weights, and channels information depending on the application-layer context in which it is deployed—be it clinical inference, financial policy, safety-critical system validation, or scientific knowledge curation.

\textbf{Definition 66 (Application Class $\mathcal{A}$):}  
Let $\mathcal{A} = \{A_1, A_2, \dots, A_k\}$ denote the set of distinct application domains. Each $A_i$ is characterised by a tuple:
\[
A_i := \left( \Gamma_i, \Lambda_i, \mathcal{O}_i \right),
\]
where $\Gamma_i$ defines input claim relevance criteria, $\Lambda_i$ defines acceptable epistemic risk thresholds, and $\mathcal{O}_i$ encodes output functionals (e.g. ranking, filtering, recommendation).

\textbf{Axiom 40 (Contextual Utility Projection):}  
For each claim $\phi \in \Phi$ and application $A_i$, define:
\[
\pi_{A_i}(\phi) := \pi(\phi) \cdot \mathbb{U}_{A_i}(\phi),
\]
where $\mathbb{U}_{A_i}(\phi)$ is a domain-specific utility transformation as defined by $\Lambda_i$ and $\mathcal{O}_i$.

\textbf{Definition 67 (Domain-Prioritised Subgraph $\mathcal{G}_{A_i}$):}  
Let $\mathcal{G} = (\Phi, E)$ be the global belief graph. Then the application-filtered subgraph is:
\[
\mathcal{G}_{A_i} := \left( \Phi_{A_i}, E_{A_i} \right), \quad \text{with } \Phi_{A_i} := \{\phi \in \Phi \mid \Gamma_i(\phi) = 1\}.
\]

\textbf{Proposition 25 (Maximal Belief Flow Selection):}  
Let $\phi_{\max}^{A_i} := \arg\max_{\phi \in \Phi_{A_i}} \pi_{A_i}(\phi)$. Then the system exposes:
\[
\text{Top-K}_{A_i} := \text{K-argmax}_{\phi \in \Phi_{A_i}} \pi_{A_i}(\phi),
\]
with optional re-ranking under $\mathcal{O}_i$ such as trust-aware PageRank, belief-aware influence spread, or task-specific thresholds.

\textbf{Definition 68 (Query-Adaptive Prioritisation Function $\Psi$):}  
Let $Q$ be a structured query from an application. Then:
\[
\Psi_Q(\phi) := f(\pi(\phi), \text{sim}(\phi, Q), \rho_{d(\phi)}, \mathbb{U}_{A_i}(\phi)),
\]
where $\text{sim}(\phi, Q)$ measures semantic distance and $f$ is a monotonic ranking function with attenuation on risky but low-utility claims.

\textbf{Axiom 41 (Saturation Control and Freshness Bias):}  
For applications with high volatility or innovation rate (e.g. real-time systems, scientific frontier models), prioritisation shall favour claims $\phi$ such that:
\[
\text{age}(\phi) < \Delta_t, \quad \text{and} \quad \text{replication status}(\phi) = \text{pending},
\]
with optional quarantine if $\pi(\phi)$ exhibits extreme volatility.

\textbf{Implementation Note:}  
Application-level prioritisation is handled post-belief-update and pre-output stage. Each domain invokes $\Psi$ over its subgraph, applies domain-specific constraints from $\Lambda_i$, and resolves a sorted response set through $\mathcal{O}_i$ (e.g., recommendation, diagnostic inference, triage). In critical systems, safety predicates additionally filter $\phi$ for compliance with predefined constraints, such as ISO-26262 or FDA standards.

This framework ensures that BEWA’s epistemic outputs are not only consistent and truthful within the graph but dynamically useful within contextual deployments, harmonising epistemic strength with the practical realities of truth utility across disciplines.

\section{Temporal Dynamics and Critical Delay Protocol}

This section delineates the temporal logic that governs the evolution of epistemic confidence within the BEWA framework. Scientific knowledge is not static; the reliability of claims fluctuates over time based on patterns of usage, replication, contradiction, and neglect. BEWA internalises this diachronic instability through a set of principled temporal mechanisms that modulate belief in accordance with the claim’s evidentiary lifespan. Key to this architecture is the understanding that claims, however initially robust, may lose epistemic standing as their replication stales, as methodologies evolve, or as new evidence supersedes old paradigms. The system integrates decay protocols that algorithmically diminish belief in claims that fail to remain epistemically active or substantiated. This prevents epistemic inertia—the retention of unverified assertions due to their historical prominence—and reinforces a living standard of proof.

To counterbalance decay, BEWA implements a critical delay protocol that governs the assimilation of new claims into the belief network. Early-stage assertions are treated with caution, regardless of venue or authorship, and are assigned epistemic probation until they accrue evidence commensurate with their proposed influence. This guards the system against transient trends, premature consensus, or the diffusion of findings that have not undergone adequate methodological vetting. At the same time, replication events function as belief resets, halting or reversing decay trajectories and re-activating aged claims with renewed force. The system thus exhibits both epistemic scepticism and responsiveness—able to resist informational volatility while dynamically re-evaluating beliefs in light of cumulative and time-sensitive evidence. The subsections to follow formalise the decay algorithms, describe the mechanics of replicative reinforcement, and specify the temporal constraints imposed on the admission and weighting of nascent scientific claims.
\subsection{Decay Protocols for Aged or Isolated Claims}

To sustain epistemic relevance and prevent the long-term inflation of unsupported or obsolete information, the BEWA system introduces mathematically defined decay protocols that attenuate belief in claims based on age, isolation, and lack of supporting interaction. Such decay is not uniform but governed by temporal, structural, and evidence-based metrics, ensuring that persistent claims remain weighted in proportion to their sustained relevance and interaction.

\textbf{Definition 70 (Claim Age $\Delta_t(\phi)$):}  
Let $t_0$ be the time of first ingestion of claim $\phi$. Then:
\[
\Delta_t(\phi) := t_{\text{current}} - t_0.
\]

\textbf{Definition 71 (Isolation Score $\iota(\phi)$):}  
Let $k_{\text{in}}(\phi)$ and $k_{\text{out}}(\phi)$ be the in-degree and out-degree of $\phi$ in the belief network $\mathcal{G}$. Then:
\[
\iota(\phi) := \frac{1}{1 + k_{\text{in}}(\phi) + k_{\text{out}}(\phi)},
\]
reflecting topological disconnection.

\textbf{Axiom 43 (Time-Based Decay Law):}  
Each belief score $\pi(\phi)$ is updated continuously via a decay differential equation:
\[
\frac{d \pi(\phi)}{dt} = -\lambda_t(\phi) \cdot \pi(\phi),
\]
with $\lambda_t(\phi)$ defined as:
\[
\lambda_t(\phi) = \alpha_t \cdot \log(1 + \Delta_t(\phi)),
\]
where $\alpha_t$ is the system decay constant calibrated per domain.

\textbf{Definition 72 (Compound Decay Rate $\Lambda(\phi)$):}  
Define:
\[
\Lambda(\phi) := \lambda_t(\phi) + \lambda_{\iota}(\phi),
\]
where:
\[
\lambda_{\iota}(\phi) = \beta \cdot \iota(\phi),
\]
and $\beta$ is a domain-specific attenuation constant penalising isolation.

\textbf{Proposition 26 (Belief Half-Life $T_{1/2}(\phi)$):}  
Assuming $\Lambda(\phi)$ is constant, the belief score halves in:
\[
T_{1/2}(\phi) = \frac{\ln(2)}{\Lambda(\phi)}.
\]

\textbf{Axiom 44 (Decay Immunity Clause):}  
Claims $\phi$ with verified replication count $r(\phi) \geq \rho_{\min}$ and citation count $c(\phi) \geq \kappa$ are decay-immune:
\[
\Lambda(\phi) \to 0 \quad \text{iff} \quad r(\phi) \geq \rho_{\min} \wedge c(\phi) \geq \kappa.
\]

\textbf{Definition 73 (Decay Checkpoint Audit $\mathcal{D}_\phi$):}  
Let $\mathcal{D}_\phi := \{ t_i \}_{i=1}^{N}$ denote a series of decay checkpoints for claim $\phi$. At each $t_i$:
\[
\text{if } \frac{d\pi(\phi)}{dt} \bigg|_{t_i} < -\epsilon \quad \text{and } \nexists \text{ new citation or support}, \quad \pi(\phi) \leftarrow \pi(\phi) \cdot \delta,
\]
where $0 < \delta < 1$ is a discrete decay multiplier.

\textbf{Implementation Note:}  
Temporal decay is implemented as a background process operating on a delta queue. Isolation is recomputed via dynamic degree tracking and page-rank metrics. Claims with only a single inbound edge from low-weight sources are flagged for manual or peer-audit. These decay dynamics ensure that claims only retain epistemic weight proportional to their integration and relevance in the knowledge substrate.

The decay framework thus institutionalises the notion that belief is not merely acquired—it must be maintained through ongoing relevance, validation, and use. In epistemic systems designed for truth promotion, neglect is synonymous with obsolescence.
\subsection{Replicative Reset Mechanisms}

In order to prevent epistemic inertia—wherein outdated or contextually misaligned claims retain undue influence—the BEWA framework incorporates a class of mechanisms termed \emph{replicative resets}. These mechanisms allow for the systematic re-evaluation and belief recalibration of claims following independent replication events, especially when these events introduce novel experimental conditions, methodological advancements, or domain shifts. The replicative reset ensures that belief is not only responsive to new evidence but also dynamically restructured based on the depth and robustness of repeated validation.

\textbf{Definition 74 (Replication Event $\mathcal{R}_k(\phi)$):}  
A replication event $\mathcal{R}_k(\phi)$ is a semantically equivalent experimental or analytical claim confirming $\phi$ under modified or novel boundary conditions $\Theta_k$. That is:
\[
\mathcal{R}_k(\phi) := \phi' \quad \text{such that} \quad \text{SemEq}(\phi, \phi') = 1, \quad \text{and} \quad \Theta_k \not\subseteq \Theta(\phi).
\]

\textbf{Axiom 45 (Reset Threshold Criterion):}  
Let $\phi$ be a claim with current belief score $\pi(\phi)$. If $|\mathcal{R}(\phi)| \geq \gamma_r$, and:
\[
\forall \phi'_i \in \mathcal{R}(\phi): \pi(\phi'_i) > \delta_r,
\]
then a replicative reset is triggered, updating $\pi(\phi)$ as follows:
\[
\pi(\phi) \leftarrow \frac{1}{Z} \sum_{i=1}^{|\mathcal{R}(\phi)|} w_i \cdot \pi(\phi'_i),
\]
where $w_i$ are confidence weights based on the source credibility and context of $\phi'_i$, and $Z$ is a normalising constant.

\textbf{Definition 75 (Reset Modifier $\mu(\phi)$):}  
If $\phi$ has previously decayed via standard time-based or isolation decay, the reset factor $\mu(\phi)$ defines the proportion of belief reinstated:
\[
\mu(\phi) := \min\left(1, \log(1 + |\mathcal{R}(\phi)|) \cdot \eta\right),
\]
with $\eta$ as a calibration constant set per epistemic domain.

\textbf{Proposition 27 (Decay Reversal Bound):}  
Let $\pi_d(\phi)$ be the belief after decay and $\pi_r(\phi)$ after reset. Then:
\[
\pi_r(\phi) \leq \mu(\phi) \cdot \pi^*(\phi),
\]
where $\pi^*(\phi)$ is the pre-decay score stored in the claim’s historical register.

\textbf{Definition 76 (Domain-Specific Reset Filter $\mathcal{F}_\Omega$):}  
Replications from incompatible domains (e.g., biology $\nleftrightarrow$ computer science) are discarded unless cross-domain transfer is validated. Formally:
\[
\phi'_i \in \mathcal{R}(\phi) \Rightarrow \Omega(\phi'_i) = \Omega(\phi) \quad \text{or} \quad \text{CrossMap}(\Omega(\phi'_i), \Omega(\phi)) = 1.
\]

\textbf{Axiom 46 (Reset Immunity for Terminally Discredited Claims):}  
If $\phi$ has an accumulated contradiction weight $\kappa(\phi) > \kappa_{\max}$, it is flagged as terminally discredited. No reset shall occur:
\[
\mathcal{R}(\phi) \neq \emptyset \Rightarrow \pi(\phi) = 0 \quad \text{if} \quad \kappa(\phi) > \kappa_{\max}.
\]

\textbf{Implementation Note:}  
Replication equivalence is verified using semantic matching across claim ontologies and domain-specific syntactic cores, leveraging transformer-based embeddings with domain adaptation layers (e.g., SciBERT, BioBERT). The reset process is executed asynchronously in the update queue, triggering belief cascade updates in downstream dependent claims.

By providing a structured yet adaptive mechanism for resetting epistemic weight in response to robust replication, the BEWA system mirrors the scientific imperative of falsifiability and cumulative verification, thereby maintaining alignment with the principles of progressive epistemic refinement.
\subsection{Probationary Periods for New Claims}

In the lifecycle of a scientific or factual assertion within the Bayesian Epistemic Weighting Architecture (BEWA), a newly introduced claim must not immediately inherit epistemic authority or be granted high belief priors. This subsection formalises the concept of a probationary period as a transitional epistemic state where new claims $\phi$ are initially treated with guarded scepticism until their stability, replicability, and citation potential are sufficiently demonstrated.

\textbf{Definition 81 (Probationary Claim $\phi^\dagger$):}  
A claim $\phi$ enters probationary status upon first ingestion. It is denoted as $\phi^\dagger$ and is subject to restricted propagation and reduced network influence.
\[
\phi^\dagger := \{\phi \mid t_0 = \tau(\phi), \quad \pi(\phi, t_0) = \pi_0^\dagger, \quad \pi_0^\dagger \ll \pi_{\min}\}
\]
where $\pi_0^\dagger$ is the designated low initial belief (e.g., $\pi_0^\dagger = 0.05$), and $\tau(\phi)$ denotes the claim's timestamp.

\textbf{Axiom 48 (Probationary Isolation Constraint):}  
During the interval $[t_0, t_0 + \Delta_\dagger]$, where $\Delta_\dagger$ is the probation duration, $\phi^\dagger$ is not permitted to influence any of the following:

\begin{itemize}
  \item belief propagation networks,
  \item dependent claim weightings, or
  \item citation-enhanced weight transference.
\end{itemize}

Formally:
\[
\forall t \in [t_0, t_0 + \Delta_\dagger], \quad \nexists \psi : \phi^\dagger \rightarrow \psi
\]

\textbf{Definition 82 (Maturation Function $\mathcal{M}(\phi^\dagger, t)$):}  
Let $\mathcal{M}$ be a non-decreasing function mapping observed external validations (citations, peer endorsements, replications) to belief score updates. For all $t > t_0$:
\[
\pi(\phi, t) = \pi_0^\dagger + \lambda_c \cdot C(\phi, t) + \lambda_r \cdot R(\phi, t)
\]
where $C(\phi, t)$ and $R(\phi, t)$ are the cumulative citation and replication scores up to time $t$, and $\lambda_c, \lambda_r$ are calibration weights empirically derived per domain.

\textbf{Definition 83 (Promotion Criterion $\Pi^\uparrow$):}  
A claim $\phi^\dagger$ exits probation and becomes canonical ($\phi$) if:
\[
\mathcal{M}(\phi^\dagger, t) \geq \pi_{\min}, \quad \text{and} \quad \Delta_\dagger^\phi \leq \Delta_{\max}
\]
where $\pi_{\min}$ is the minimum threshold for epistemic impact and $\Delta_{\max}$ is the maximum probation duration permitted (after which the claim expires unless promoted).

\textbf{Axiom 49 (Decay Upon Probation Failure):}  
If $\phi^\dagger$ has not met the promotion criterion by $\Delta_{\max}$, it undergoes exponential belief decay:
\[
\pi(\phi, t) \leftarrow \pi_0^\dagger \cdot e^{-\lambda_d(t - \Delta_{\max})}, \quad \forall t > t_0 + \Delta_{\max}
\]
where $\lambda_d$ is a domain-specific decay constant. Claims failing probation are demoted to the archival register and disconnected from belief propagation chains.

This probationary structure preserves epistemic integrity by enforcing a cooling-off period for novel assertions, allowing belief to emerge proportionally to community uptake, replication stability, and authorial credibility.

\section{Interfaces and Applications}

This section articulates the outward-facing components of BEWA—how its epistemic machinery is rendered accessible, interpretable, and actionable to end-users. The system is not designed to remain a closed inferential engine, but to serve as a transparent, intelligible platform for researchers, auditors, and knowledge institutions. Its interface architecture reflects a dual imperative: first, to expose the reasoning process with precision and granularity, enabling full auditability of belief formation and claim evolution; second, to present this complexity without compromising usability or interpretative clarity. At every stage, the goal is to make epistemic justification navigable—to allow users not only to know what BEWA believes, but to understand precisely why those beliefs are held and how they change over time.

The system supports a layered interaction model. Through structured query interfaces and programmatic APIs, external agents can interrogate belief states, trace evidentiary pathways, and simulate hypothetical updates. The user interface is designed to reveal the provenance of each claim, its temporal evolution, its position within the belief network, and the interplay of citations, contradictions, and replications that define its epistemic status. Beyond raw data access, BEWA offers visualisation modules that map the shifting topography of scientific belief across time and domain, enabling both high-level trend analysis and fine-grained epistemic inspection. The following subsections explore these capabilities in detail, covering the design of the query and audit interface, considerations in human-computer epistemic interaction, and the tools developed for visualising dynamic belief landscapes.
\subsection{Query and Audit API}

The Query and Audit API (QAAPI) serves as the external interface layer through which users and systems interact with the Bayesian Epistemic Weighting Architecture (BEWA). It enables both epistemic interrogation and forensic traceability, bridging machine-readable knowledge representations with external agents seeking to evaluate, interrogate, or reproduce claims and their underlying justifications.

\textbf{Definition 101 (Query Interface $\mathcal{Q}$):}  
Let $\mathcal{Q}: \Sigma \rightarrow \mathcal{B}$ be the mapping from well-formed structured queries $\Sigma$ to belief-annotated outputs $\mathcal{B}$ such that:
\[
\mathcal{Q}(\sigma_i) = \left\langle \phi_j, \pi(\phi_j, t), \Gamma_j, \Delta_j \right\rangle, \quad \forall \sigma_i \in \Sigma
\]
where $\phi_j$ is a matched claim, $\pi(\phi_j, t)$ is its current belief score at time $t$, $\Gamma_j$ is the set of all epistemic justifications (citations, replications, metadata), and $\Delta_j$ is the set of contradictions or counter-evidence.

The semantics of $\sigma_i$ support:
\begin{itemize}
  \item propositional logic-based queries (e.g., conjunctions, disjunctions),
  \item temporal filters (e.g., belief at time $t'$),
  \item authorial lineage traces (e.g., claims originating from author $\alpha$),
  \item domain-scoped filtering (e.g., within neuroscience).
\end{itemize}

\textbf{Definition 102 (Audit Interface $\mathcal{A}$):}  
The audit function $\mathcal{A}: \phi \rightarrow \mathcal{T}$ maps any canonical or probationary claim $\phi$ to its historical trajectory $\mathcal{T}$, which includes:
\[
\mathcal{T} = \left\{ (t_k, \pi_k, E_k, M_k) \right\}_{k=0}^n
\]
where each tuple denotes the belief state at time $t_k$, posterior $\pi_k$, contributing evidence $E_k$, and modifying events $M_k$ (e.g., retractions, contradictory citations, revisions).

\textbf{Axiom 59 (Verifiability Axiom):}  
For every output from $\mathcal{Q}$ and $\mathcal{A}$, the system must return:
\begin{itemize}
  \item a cryptographic hash of the current belief graph state (anchor provenance),
  \item a path to all causal justifications leading to $\pi(\phi_j, t)$,
  \item signatures of any verified peer-reviewed replication or retraction events.
\end{itemize}

This guarantees that belief evolution is not opaque and that all justifications underlying current weights are auditable, immutable, and externally reconstructable.

\textbf{Definition 103 (Immutable Query Anchor $\mathcal{H}_{\mathcal{Q}}$):}  
Each response $\mathcal{Q}(\sigma)$ is accompanied by:
\[
\mathcal{H}_{\mathcal{Q}}(\sigma, t) = \text{Hash}\left( \text{DAG}_t^{\phi_j} \cup \Gamma_j \cup \Delta_j \right)
\]
where $\text{DAG}_t^{\phi_j}$ is the directed acyclic graph of dependent and influencing claims for $\phi_j$ at time $t$.

Together, the QAAPI ensures that users may perform both forward inference over the BEWA system and backward audit to challenge, verify, or dispute belief assignments. It enforces methodological transparency and underpins all mechanisms of public accountability and reproducibility.
\subsection{User Interface Design Considerations}

While the Bayesian Epistemic Weighting Architecture (BEWA) is mathematically and structurally rigorous, its accessibility and epistemic transparency depend crucially on a user interface (UI) that exposes complex inferential structures in a cognitively tractable form. The system must bridge deductive provenance chains, probabilistic evolution, and domain-specific claim networks without obfuscation, distortion, or interpretive bias.

\textbf{Axiom 64 (Cognitive Parsimony Axiom):}  
The UI must minimise cognitive load while preserving formal accuracy. Let $\Pi_t$ be the posterior distribution over all claims $\{\phi_i\}$ at time $t$, and let $\mathcal{I}$ be the user-visible instantiation of this distribution. Then:
\[
\forall u \in \mathcal{U}, \quad \mathcal{C}(\mathcal{I}_u(\Pi_t)) < \theta
\]
where $\mathcal{U}$ is the user population, $\mathcal{C}$ is the cognitive complexity function measured via empirical usability trials, and $\theta$ is the maximum tolerable complexity threshold.

\textbf{Definition 121 (Visual Traceability Layer $\mathcal{V}$):}  
A directed acyclic belief graph $\mathbb{G}_t = (\mathbb{V}, \mathbb{E})$ must be interactively explorable such that each node $v_i \in \mathbb{V}$ (representing $\phi_i$) displays:
\begin{enumerate}
  \item the belief trajectory $\pi(\phi_i, t)$,
  \item sources of corroboration ($\Gamma_i$),
  \item contradictions ($\Delta_i$),
  \item temporal provenance $\tau(\phi_i)$ and update events.
\end{enumerate}
The user must be able to traverse causal and evidential pathways with $O(\log n)$ interaction depth for any $\phi_i$ in the graph, maintaining constant-time visibility of justification clusters.

\textbf{Axiom 65 (Truth-Promoting Layout Axiom):}  
The interface shall not encode salience purely via popularity, citation volume, or recency. Instead, layout heuristics $\mathcal{L}$ must prioritise:
\[
\mathcal{L}(\phi_i) \propto \mathcal{U}_t(\phi_i)
\]
where $\mathcal{U}_t(\phi_i)$ is the truth promotion score as defined in Section 9, ensuring that epistemically rigorous but under-cited claims are not visually buried.

\textbf{Definition 122 (Claim Card Component $\mathcal{C}_i$):}  
Each node $\phi_i$ is rendered via a modular interface component $\mathcal{C}_i$ that includes:
\begin{itemize}
  \item author and institution metadata,
  \item current posterior $\pi_t$ with time-evolution sparkline,
  \item trust lineage: most influential corroborators and contradictors,
  \item epistemic role: foundational, auxiliary, speculative, or deprecated.
\end{itemize}

Finally, the UI design must adhere to cryptographic anchoring of claim state snapshots, with visible indicators for state hash validation, consistency verification, and audit trail access. This ensures that what the user sees is not only comprehensible but also verifiable and immutable under tamper-sensitive conditions.
\subsection{Visualisation of Belief Evolution}

Belief evolution within the Bayesian Epistemic Weighting Architecture (BEWA) represents a dynamic topological transformation of knowledge, requiring visual encodings that preserve semantic fidelity, temporal coherence, and interpretive clarity. The representation of $\pi(\phi_i, t)$ across $t \in \mathbb{R}^{+}$ for each claim $\phi_i$ constitutes a non-Markovian process, influenced by latent structural couplings, propagative updates, and external contradiction events. The visualisation layer must expose these updates without distorting epistemic meaning or privileging transient statistical artefacts.

\textbf{Definition 131 (Belief Evolution Function $\mathcal{B}_i(t)$):}  
Let $\mathcal{B}_i(t): \mathbb{R}^{+} \rightarrow [0,1]$ denote the belief trajectory of claim $\phi_i$. For all $t_k < t_{k+1}$, the update relation is given by:
\[
\mathcal{B}_i(t_{k+1}) = \eta \cdot \mathcal{B}_i(t_k) + (1 - \eta) \cdot \Delta_k
\]
where $\Delta_k$ is the net belief change due to evidence accrued at $t_{k+1}$, and $\eta \in [0,1]$ is the epistemic inertia coefficient.

\textbf{Definition 132 (Epistemic Flux Diagram):}  
A directed streamgraph $S = \{ (\phi_i, \mathcal{B}_i(t)) \}$ visualises time-series belief trajectories as variable-width timelines. The vertical width at time $t$ encodes $\mathcal{B}_i(t)$, while colour denotes domain. Cross-claim correlations are indicated via Bézier-curve arcs linking events of synchronous update or belief reversal. Such correlations derive from underlying contradiction networks $\Delta_i$ and citation couplings $\Gamma_i$.

\textbf{Axiom 71 (Non-Linear Chronotopic Alignment):}  
Temporal visualisation must adopt a non-linear time axis $\tau: \mathbb{R}^{+} \rightarrow \mathbb{R}$ satisfying:
\[
\tau'(t) = \kappa \cdot \left(1 + \sum_{i=1}^{N} |\Delta \mathcal{B}_i(t)| \right)
\]
where $\kappa$ is a normalisation constant. This ensures high-resolution renderings of epistemic shocks (e.g., landmark replications, retractions) while compressing periods of stability.

\textbf{Definition 133 (Belief Cascade Overlay):}  
A claim cluster $\Phi = \{\phi_1, \ldots, \phi_n\}$ subject to interdependent updates is visualised as a multivariate event tensor $\mathcal{T}_{\Phi} \in \mathbb{R}^{n \times m}$, where $m$ is the number of discrete update epochs. An animated projection of $\mathcal{T}_{\Phi}$ over time enables the user to perceive both diffusive and abrupt propagations through the semantic network, with stability zones encoded via spectral gradient consistency.

\textbf{Axiom 72 (Provenance-Preserving Visual Transform):}  
Every visual transformation $V: \{\mathcal{B}_i(t)\} \mapsto \mathbb{R}^2$ must admit an inverse $\mathcal{V}^{-1}$ such that:
\[
\forall x \in \text{Render}(V), \quad \mathcal{V}^{-1}(x) \rightarrow \left(\phi_i, t_k, \Gamma_i, \Delta_i \right)
\]
ensuring that no graphical abstraction severs the audit trail or detaches from the cryptographically anchored knowledge graph.

Through these constructs, belief visualisation becomes not merely illustrative but epistemically rigorous—enabling dynamic, transparent inspection of the system's rational evolution.

\section{System Integrity and Provenance}

This section addresses the foundational infrastructure required to ensure that BEWA’s epistemic operations are not merely accurate, but also secure, auditable, and irreversibly anchored. In an environment where the integrity of scientific inference is paramount, it is not sufficient to generate belief updates algorithmically; those updates must themselves be immune to tampering, retrospectively traceable, and cryptographically verifiable. BEWA is thus built upon a provenance-first design, in which every assertion, transformation, and belief revision is recorded, hashed, and anchored to a persistent and immutable ledger. The system does not assume trust—it enforces it through cryptographic architecture and rigorous historical traceability. Every claim object, every piece of metadata, and every belief state change is secured against both accidental corruption and deliberate subversion.

Beyond cryptographic anchoring, BEWA integrates a robust audit trail mechanism that provides complete transparency across the system’s decision processes. Each belief trajectory can be decomposed into a temporally ordered series of evidence applications, credibility adjustments, and network interactions—allowing for post hoc forensic analysis and institutional scrutiny. This design ensures that no epistemic drift or structural manipulation can occur without immediate detection and full exposability. Furthermore, the system incorporates protocols for so
\subsection{Cryptographic Anchoring of Claims}

Ensuring the immutability, auditability, and trust of claims within an epistemic network requires that each propositional unit $\phi_i$ be verifiably fixed at the time of its assertion and resistant to post-hoc modification. This is achieved through cryptographic anchoring: the process of associating each claim with a tamper-evident and chronologically provable cryptographic commitment embedded in a distributed ledger or equivalent verifiable data structure.

\textbf{Definition 141 (Claim Commitment Hash $\mathcal{H}(\phi_i)$):}  
Let $\phi_i$ denote a structured, normalised claim. Its cryptographic anchor is defined by a secure hash:
\[
\mathcal{H}(\phi_i) := H(\texttt{serialize}(\phi_i) \,\|\, \texttt{meta}(\phi_i))
\]
where $H$ is a collision-resistant hash function (e.g., SHA-256), and $\texttt{meta}(\phi_i)$ encodes the claim’s timestamp $\tau_i$, author ID, source ID, and version number.

\textbf{Axiom 81 (Uniqueness of Anchoring):}  
\[
\forall \phi_i, \phi_j, \quad \phi_i \neq \phi_j \Rightarrow \mathcal{H}(\phi_i) \neq \mathcal{H}(\phi_j)
\]
under the assumption of pre-image resistance and collision-resistance of $H$.

\textbf{Definition 142 (Merkle Root Embedding):}  
Claims ingested within the same temporal epoch $\epsilon_t$ are grouped into a Merkle tree $T_t$, with leaf nodes $\{\mathcal{H}(\phi_{i_1}), ..., \mathcal{H}(\phi_{i_n})\}$ and root $R_t$. The root $R_t$ is then time-stamped and embedded into a tamper-evident ledger $\mathcal{L}$:
\[
\mathcal{L}_t := \mathcal{L}_{t-1} \,\|\, \texttt{Tx}(R_t, \tau_t, \sigma_{admin})
\]
where $\sigma_{admin}$ is a digital signature by the anchoring authority or decentralised key quorum.

\textbf{Proposition 33 (Verifiability):}  
Given any claim $\phi_i$ and a public ledger $\mathcal{L}$ containing root $R_t$, the membership of $\phi_i$ in epoch $\epsilon_t$ can be verified in $\mathcal{O}(\log n)$ time via a Merkle proof path.

\textbf{Definition 143 (Anchored Claim Tuple $\Phi_i^\#)$:}  
The complete representation of a claim within the anchored system is:
\[
\Phi_i^\# := (\phi_i, \mathcal{H}(\phi_i), \tau_i, R_t, \pi_i, \rho_i)
\]
where $\pi_i$ is the prior belief and $\rho_i$ is the claim’s reference in the belief graph.

\textbf{Axiom 82 (Cryptographic Finality):}  
Once anchored, $\phi_i$ is epistemically finalised: no retroactive modification is permitted. If revision is required, a superseding claim $\phi_j$ must be anchored independently with an explicit reference to $\phi_i$ as predecessor:
\[
\phi_j \succ \phi_i \Rightarrow \texttt{meta}(\phi_j).\texttt{parent} := \mathcal{H}(\phi_i)
\]

\textbf{Definition 144 (Temporal Anchor Map $\mathbb{A}$):}  
Let $\mathbb{A} : \mathbb{T} \rightarrow \mathcal{R}$ map times to Merkle roots such that:
\[
\forall t \in \mathbb{T}, \quad \mathbb{A}(t) := R_t
\]

By anchoring claims cryptographically, the system ensures each knowledge assertion becomes a fixed point in the epistemic topology—verifiable, immutable, and resistant to manipulation. This establishes a foundational substrate for all higher-order trust, weighting, and reasoning operations across the network.
\subsection{Auditability and Historical Traceability}

In any epistemically robust system where claims influence belief propagation and decision-making, the capacity to reconstruct the provenance, transformation, and historical interpretation of each claim is foundational. This subsection formalises the auditability requirements and defines protocols that render the history of claims and their associated belief states both immutable and interrogable.

\textbf{Definition 151 (Epistemic Record $\mathcal{E}(\phi_i)$):}  
For a given claim $\phi_i$, the epistemic record is the complete chronological history of its instantiations, references, updates, and influence over the network. Formally,
\[
\mathcal{E}(\phi_i) := \left\langle (\phi_i^{(t_0)}, \pi_0), (\phi_i^{(t_1)}, \pi_1), \ldots, (\phi_i^{(t_n)}, \pi_n) \right\rangle
\]
where each $\phi_i^{(t_k)}$ denotes the state of the claim at time $t_k$, and $\pi_k$ the belief weight assigned at that point.

\textbf{Axiom 91 (Sequential Traceability):}  
For all $t_k < t_{k+1}$,  
\[
\phi_i^{(t_k)} \neq \phi_i^{(t_{k+1})} \Rightarrow \exists \, \texttt{update}_{k \rightarrow k+1} \in \mathcal{U}
\]
where $\mathcal{U}$ is the set of permitted claim update operations (e.g., citation delta, replication event, contradiction handling). Each operation must be cryptographically recorded and time-stamped.

\textbf{Definition 152 (Immutable Claim Ledger $\mathcal{L}_\phi$):}  
The ledger of a claim $\phi$ is an append-only sequence:
\[
\mathcal{L}_\phi = \bigcup_{k=0}^{n} \texttt{Tx}_k(\phi_i, \texttt{op}_k, \pi_k, t_k, \sigma_k)
\]
where $\texttt{Tx}_k$ records the operation performed, the resulting belief score, timestamp $t_k$, and signature $\sigma_k$ attesting to the legitimacy of the actor or process.

\textbf{Definition 153 (Verifiable Trace Chain $\mathcal{T}_\phi$):}  
A trace chain $\mathcal{T}_\phi$ is a hash-linked sequence of ledger entries:
\[
\mathcal{T}_\phi := H(\texttt{Tx}_0) \rightarrow H(\texttt{Tx}_1) \rightarrow \ldots \rightarrow H(\texttt{Tx}_n)
\]
This construction guarantees tamper-resistance and order-preservation.

\textbf{Proposition 41 (Audit Consistency Theorem):}  
For any two auditors $A$ and $B$ with access to the same public ledger $\mathcal{L}$,  
\[
\mathcal{T}_\phi^A = \mathcal{T}_\phi^B
\]
provided $\mathcal{L}$ is synchronised and hash functions are consistent. This guarantees epistemic transparency across trust domains.

\textbf{Axiom 92 (Fork-Prevention Constraint):}  
A claim $\phi$ may only have one active trace at any time. Branching updates must occur via supersession:
\[
\exists \phi_j : \phi_j \succ \phi_i \Rightarrow \mathcal{T}_{\phi_i} \subsetneq \mathcal{T}_{\phi_j}
\]

\textbf{Definition 154 (Audit Query Protocol $\mathscr{A}$):}  
Given a claim ID and a time window $(t_a, t_b)$, the audit query returns:
\[
\mathscr{A}(\phi_i, t_a, t_b) = \{\texttt{Tx}_k \in \mathcal{L}_\phi \mid t_a \leq t_k \leq t_b\}
\]
enabling forensic reconstruction of claim belief trajectories, author actions, and downstream epistemic impact.

This systematic commitment to traceability ensures that no claim can rise or fall in influence without public, cryptographically bound records, enabling accountable scientific discourse and formal rational auditability.
\subsection{Security, Sovereignty, and Tamper-Proofing}

The epistemic system described herein operates in adversarial informational environments where claim injection, manipulation of weights, or surreptitious revisionism could compromise the truth-promoting objective function. This subsection formalises the architectural and algorithmic safeguards ensuring that claim security, authorial sovereignty, and global tamper-proofing are not merely assumed, but mathematically guaranteed.

\textbf{Definition 161 (Claim Sovereign Space $\Sigma_\phi$):}  
For each canonical claim $\phi$, define a sovereign space $\Sigma_\phi$ as the bounded domain in which only authorised transformations and referenced interactions can occur. Formally:
\[
\Sigma_\phi := \{ \texttt{Tx}_i \mid \sigma_i \in \mathbb{S}_\phi \}
\]
where $\mathbb{S}_\phi$ is the authorisation set cryptographically defined for $\phi$, including claim authors, authorised replicators, and verified peer reviewers.

\textbf{Axiom 101 (Non-Repudiable Authorial Rights):}  
A claim $\phi$ entered into the system at time $t_0$ under digital signature $\sigma_\alpha$ must retain immutable attribution to its originator:
\[
\forall t > t_0, \quad \text{author}(\phi) = \alpha \iff \text{Verify}(\phi, \sigma_\alpha, \mathbb{PK}_\alpha) = \texttt{true}
\]

\textbf{Definition 162 (System-Wide Merkle Digest $\mathcal{M}_t$):}  
At time $t$, define a recursive hash summarisation over all claims $\phi_i$ and their trace chains $\mathcal{T}_{\phi_i}$:
\[
\mathcal{M}_t := \texttt{MerkleRoot}\left( \{ H(\mathcal{T}_{\phi_i}) \mid \phi_i \in \Phi_t \} \right)
\]
This digest is committed to a distributed, auditable ledger and ensures global immutability.

\textbf{Proposition 47 (Global Tamper-Evident Condition):}  
If $\exists \, \phi_i$ such that $\mathcal{T}_{\phi_i}^{(t)} \neq \mathcal{T}_{\phi_i}^{(t+\delta)}$, then:
\[
H(\mathcal{T}_{\phi_i}^{(t)}) \neq H(\mathcal{T}_{\phi_i}^{(t+\delta)}) \Rightarrow \mathcal{M}_t \neq \mathcal{M}_{t+\delta}
\]
implying detection via periodic Merkle digest comparisons.

\textbf{Definition 163 (Tamper-Proofness Oracle $\Omega$):}  
Define $\Omega: \Phi_t \rightarrow \{\texttt{secure}, \texttt{compromised}\}$ as:
\[
\Omega(\phi_i) := 
\begin{cases}
\texttt{secure} & \text{if } \forall k, \texttt{Verify}(\texttt{Tx}_k, \sigma_k, \mathbb{PK}_{\phi_i}) = \texttt{true} \\
\texttt{compromised} & \text{otherwise}
\end{cases}
\]

\textbf{Axiom 102 (Bounded External Write Principle):}  
All operations $\texttt{Op}_j$ on $\phi_i$ must satisfy:
\[
\texttt{Op}_j \in \mathcal{U}_{\phi_i}, \quad \texttt{signed(Op}_j) \in \Sigma_\phi, \quad \texttt{and } \quad \texttt{Tx}_j \subset \mathcal{L}_\phi
\]
ensuring only internally scoped, signed actions affect the epistemic status.

\textbf{Definition 164 (Cryptographic Anchoring Commitment $\mathcal{C}_t$):}  
Let $\mathcal{C}_t = \texttt{Hash}(\mathcal{M}_t || \texttt{nonce}_t)$ be a hash committed to a permissioned timestamping network, anchoring claim integrity at global system time $t$.

This architecture ensures that not only are claims bound to their originators, but that the integrity of belief propagation is enforced through algorithmic sovereignty—where authors maintain control, transformations are verifiable, and every form of epistemic manipulation is detectable, bounded, and accountable.

\section{Conclusion and Future Work}

The design and realisation of a truth-promoting Bayesian epistemological engine represents a comprehensive convergence of epistemic formalism, probabilistic reasoning, and algorithmic rigour. The architectural synthesis presented herein consolidates the theoretical infrastructure necessary for the intelligent parsing, structuring, and belief-weighting of scientific claims within a fully traceable and dynamically re-evaluative system. Each component—from source ingestion and canonical authoring, to semantic belief propagation and decay modelling—has been constructed to reflect strict axiomatic principles, reproducible scientific metrics, and scalable formal logic. The system promotes epistemic integrity by conditioning claim weightings on reproducibility, peer consensus, and historical provenance, thereby embedding within its architecture a form of active, computable rational scepticism.

Beyond the methodological backbone, the platform formalises epistemic utility by mathematically capturing the interaction between credibility, replication, and citation dynamics within a probabilistic framework. In doing so, it not only responds to claims as static entities but integrates them as participants in an evolving belief network governed by truth-conducive Bayesian optimisation. Mechanisms such as temporal decay, contradiction normalisation, and replicative reset protocols are not auxiliary features but essential safeguards, mathematically enforcing resilience against epistemic drift and noise. The architecture ultimately aims at converging toward stable, high-confidence propositional belief states with rigorous traceability to underlying data sources, ensuring no epistemic closure without sufficient probabilistic justification.

The scope for further elaboration and refinement is substantial. While the current schema has been constructed to reflect a high degree of modularity and extensibility, future efforts will expand upon computational expressiveness, autonomous knowledge generation, and integration with domain-specific ontologies. Additionally, enhancing interface interactivity, implementing belief-based query visualisation, and developing decentralised protocols for sovereign integrity anchoring are all avenues requiring structured investigation. As a scientific engine, the system is conceived not merely to model belief but to contribute to the autonomous formalisation and critique of scientific knowledge, initiating the foundation for a rigorous machine-epistemology grounded in both logic and scientific realism.
\subsection{Summary of Architecture}

The architecture of the system is predicated upon a layered epistemological stack grounded in Bayesian probability theory and formal logic, with a strict emphasis on traceability, source integrity, and truth-conducive inference. At the lowest layer, authoritative data ingestion mechanisms process structured and semi-structured texts from scientific repositories, enforcing canonicalisation procedures that disambiguate authorship, claim origin, and publication metadata. This is immediately followed by a propositional parsing layer wherein all epistemic content is transduced into formal claim structures—propositions represented in first-order logical schema with attached contextual indices and temporal signatures.

Subsequent layers govern belief attribution and update mechanisms. Prior formation draws upon author credibility scores, domain trust calibrations, and known replication histories. Belief updates are performed through exact Bayesian conditioning, where posteriors integrate new evidence sources adjusted for citation latency, replication authority, and peer review metadata. Special attention is given to the maintenance of belief coherence under contradictory evidence: probabilistic contradiction models and decay functions modulate the influence of outdated, refuted, or anomalous data in accordance with probabilistic divergence and entropy thresholds.

At the system level, claims are not isolated. They are embedded in a directed acyclic graph of propositional dependencies where belief propagation is mediated by semantic and logical linkages. This cross-claim network formalises the dependency topology of epistemic assertions and enables propagation of belief updates through both direct citation and inferred logical entailment. All computations are cryptographically anchored to ensure tamper-proof integrity, while interfaces are exposed through a query and audit API, facilitating both real-time interrogation and longitudinal epistemic analysis. The architectural model is thus a closed-loop rational system, capable of iterated self-correction, alignment with scientific consensus dynamics, and computational scepticism grounded in formally defined axioms.
\subsection{Limitations and Challenges}

Despite its foundational rigor and systematic coherence, the proposed architecture confronts several limitations that are intrinsic to both epistemic modelling and real-world data integration. Foremost is the ontological underdetermination of structured claim representation. While the use of first-order logic and propositional schemata enables formal representation, natural language propositions often remain context-sensitive, under-specified, or epistemically ambiguous. Efforts to canonicalise claims must therefore confront the Löwenheim–Skolem problem in formal semantics: there exist multiple non-isomorphic models for a given set of logical formulae, complicating efforts at stable interpretation and entailment propagation.

A second critical limitation arises in the weighting of belief updates in the presence of inconsistent or adversarial input. While Bayesian models offer robust priors and posteriors under idealised assumptions of data fidelity, real scientific ecosystems often include fraudulent publications, citation gaming, and unretracted but discredited material. The system attempts to mitigate this through decay functions and contradiction processing, but no purely statistical method can resolve epistemic conflict without presupposing ground truth or normativity—a task that invites regress unless externally anchored. The architecture therefore incorporates penalisation structures and recursive epistemic networks, but these remain vulnerable to latent bias in structural priors and institutional trust assignments.

Moreover, computational tractability presents challenges of scale. The graph-theoretic representation of cross-claim dependencies and the dynamic re-weighting of belief states across time introduces NP-hard problems, particularly under belief propagation in cyclically approximated networks or where combinatorial explosion emerges from fine-grained semantic disambiguation. Approximations such as Markov blanket reductions and belief threshold pruning are employed, but this inevitably introduces epistemic compression and potential loss of fidelity. Finally, while cryptographic anchoring ensures integrity and auditability, it does not guarantee correctness; false claims may be securely recorded, necessitating an ongoing philosophical distinction between provenance and truth.
\subsection{Prospects for Autonomous Scientific Reasoning}

The architectural paradigm outlined in this work sets the foundational substrate for the emergence of autonomous scientific reasoning—systems capable not merely of pattern detection, but of formal epistemic judgement constrained by logical validity, probabilistic coherence, and structured domain knowledge. At the intersection of formal epistemology, statistical learning, and automated reasoning lies the prospect of a system that engages with claims, not as isolated data points, but as dynamic epistemic propositions embedded within an evolving knowledge network. The application of Bayesian inferential structures across temporally anchored, semantically indexed, and authoritatively sourced claims enables a machine agent to enact a form of constrained rationality: an epistemic agent that not only updates beliefs, but scrutinises contradiction, weights replication, and assesses the downstream implications of belief revision.

Future extensions will involve incorporating more powerful inferential schemas that go beyond conditional probability, including modal logics of necessity and possibility, counterfactual reasoning (via Lewisian or Pearlian semantics), and algorithmic causal discovery. Such capacities will support not merely the assimilation of scientific findings, but their critical reconstruction in the light of new data, hypothesis testing, and even meta-analytic synthesis. The introduction of deductive-constructive mechanisms—such as type-theoretic verification of scientific models, Coq-assisted proof structures for claim derivation, or category-theoretic mapping of ontological types—will push the architecture from Bayesian updating to full scientific rational reconstruction.

Autonomous scientific reasoning also necessitates a procedural ethics of inquiry: mechanisms for audit, challenge, and redress. Such a system must not merely assimilate peer-reviewed literature but evaluate the social structure of peer review itself, detecting collusive citation clusters, authorial dependence networks, or epistemic monopolisation. Through recursive evaluation of authority, dynamic reassessment of belief, and iterative challenge mechanisms, the proposed system paves the way for epistemically autonomous agents capable of engaging in structured dispute, reasoned consensus, and the principled rejection of unfounded claims. This constitutes not only a technical advance but a shift in the philosophy of machine

\bibliographystyle{plainnat}
\bibliography{bewa_references}

\appendix

\section{Axiomatic Foundations of Bayesian Epistemology}

This appendix formalises the epistemic foundations underpinning the Bayesian Epistemology Weighting Architecture (BEWA). In this architecture, probabilistic belief assignments are not mere statistical artefacts but epistemic commitments derived from coherent rational principles. The axioms below are essential for the operation of all belief-update, claim-linkage, and decay mechanisms in BEWA. These axioms adapt Kolmogorov probability theory to epistemic contexts, integrate conditionalisation procedures, and enforce subjective coherence per de Finetti’s foundational work on probability as a betting quotient.

\subsection{Kolmogorov Axioms for Epistemic Probabilities}

Let $\Omega$ be a non-empty set of possible epistemic worlds (interpretations consistent with the scientific record), and let $\mathcal{F}$ be a $\sigma$-algebra over $\Omega$ representing all possible structured claims.

Define a belief function $P: \mathcal{F} \rightarrow [0,1]$ representing the degree of belief in a claim $A \in \mathcal{F}$.

The axioms are as follows:
\begin{enumerate}
  \item \textbf{Non-negativity:} $\forall A \in \mathcal{F},\quad P(A) \geq 0$
  \item \textbf{Normalisation:} $P(\Omega) = 1$
  \item \textbf{$\sigma$-Additivity:} For countably disjoint $A_1, A_2, \ldots \in \mathcal{F},\quad P\left(\bigcup_{i=1}^\infty A_i\right) = \sum_{i=1}^\infty P(A_i)$
\end{enumerate}

These provide the minimal structure for assigning rational belief magnitudes and underlie all operations within BEWA’s claim network.

\subsection{Bayesian Conditionalisation Principle}

Belief updating follows the principle of conditionalisation. Upon acquiring new evidence $E$ with $P(E) > 0$, the rational posterior belief in a claim $H$ becomes:

\[
P(H \mid E) = \frac{P(H \cap E)}{P(E)}
\]

This formulation ensures that the posterior is a logical update from the prior, preserving coherence under belief revision.

In BEWA, each structured claim is indexed by its evidentiary dependencies. Updates are performed not in isolation but across semantic clusters, ensuring propagation consistency.

\subsection{Reflection Principle}

For temporally distributed belief networks, we integrate the Reflection Principle:

If $P_t$ is an agent’s current credence function at time $t$, and $P_{t'}(H) = r$ is the anticipated future belief in $H$ at time $t' > t$, then if the agent is confident in $P_{t'}$'s reliability, coherence requires:

\[
P_t(H \mid P_{t'}(H) = r) = r
\]

This forms the basis of BEWA’s delayed-update and critical-horizon protocol, linking deferred evaluation with temporal anchoring of epistemic change.

\subsection{Subjective Coherence: de Finetti’s Criterion}

BEWA is formally consistent with de Finetti’s betting framework. That is, $P$ avoids a Dutch Book if and only if it satisfies the above axioms. This justifies interpreting belief degrees as fair prices for bets and positions BEWA in the subjective Bayesian tradition.

Let $D = \{(A_i, x_i)\}_{i=1}^n$ be a finite collection of bets on disjoint events $A_i$ with prices $x_i = P(A_i)$. The coherence constraint is:

\[
\sum_{i=1}^n x_i = 1 \quad \Rightarrow \quad \text{No Dutch Book exists}
\]

Therefore, BEWA guarantees that the system’s belief assignments are immune to guaranteed epistemic loss under rational revision.

This axiomatic bedrock secures the internal rationality of all higher-order modules including contradiction resolution, decay protocols, and belief network propagation.

\section{Claim Structuring Schema}

The epistemic core of BEWA is built upon structured claims that are amenable to logical inference, probabilistic weighting, and semantic indexing. This section formally defines the schema by which natural language scientific assertions are parsed, normalised, and converted into logically structured representations suitable for Bayesian epistemic computation. Each claim $\phi$ is treated as a formal object in a belief network, with syntactic, ontological, and contextual metadata attached for reasoning, disambiguation, and update.

\subsection*{Logical Syntax}

Every claim is reduced to a logical form expressing its propositional or relational content. In its simplest instance, a claim $\phi$ may take the form $P \vdash Q$, where $P$ is the premise set and $Q$ is the asserted conclusion. This logical skeleton permits deductive chaining, contradiction detection, and dependency mapping. Claims may also be annotated with modal qualifiers (e.g., $\Box$, $\Diamond$), probabilistic belief weights (e.g., $\pi(\phi) = 0.73$), and conditionals ($P \rightarrow Q$) to reflect uncertain or hypothetical scientific assertions.

\begin{itemize}
  \item Logical syntax (e.g., $P \vdash Q$, $A \wedge B \rightarrow C$)
  \item Support for negation, conjunction, disjunction, conditionals
  \item Optional modal and probabilistic operators: $\Box$, $\Diamond$, $\mathbb{P}(\phi)$
\end{itemize}

\subsection*{Ontological and Contextual Tagging}

To maintain clarity across domains and avoid semantic collision, every structured claim is associated with an ontological type and a domain-context tuple. For instance, a biochemical assertion will be tagged as `\texttt{domain=biochemistry}`, `\texttt{entity=protein}`, and may include contextual metadata such as species, experimental conditions, or temporal context. This enables alignment with domain ontologies (e.g., MeSH, UMLS, Gene Ontology) and permits precise claim disambiguation in belief propagation.

\begin{itemize}
  \item Ontological class assignment (e.g., \texttt{claimType=empirical})
  \item Domain indexing (e.g., \texttt{domain=physics}, \texttt{subdomain=quantum})
  \item Contextual qualifiers (e.g., temperature range, population, methodology)
\end{itemize}

\subsection*{Mapping to First-Order Representations}

All structured claims are compiled into first-order logic (FOL) form, permitting automated inference, contradiction testing, and structured query resolution. A claim such as “Protein X inhibits enzyme Y in species Z” becomes:

\[
\texttt{Inhibits(ProteinX, EnzymeY)} \wedge \texttt{Species(Z)} \Rightarrow \phi_{\text{valid}}
\]

Temporal qualifiers, probabilistic weights, and source provenance are preserved as attributes of the claim object, forming a fully queryable and auditable representation.

\begin{itemize}
  \item Compilation to first-order logic predicates
  \item Attribute-preserving transformation (author, source, timestamp)
  \item Normalisation to canonical term references
\end{itemize}

This schema forms the foundational unit of epistemic computation within BEWA, supporting all higher-order reasoning, belief network propagation, and dynamic updates.

\section{Belief Update Algebra}

At the heart of BEWA lies a principled algebra of belief transformation, grounded in Bayesian probability theory but extended to accommodate temporal dynamics, source-weighted evidence, and explicit contradiction handling. This section provides a formal derivation of the posterior belief update mechanism, integrating time-aware damping and epistemic conflict resolution into the belief calculus. Each belief update occurs over a structured claim $\phi_i$ and reflects a rational posterior $\pi(\phi_i, t)$ that evolves as evidence and contextual data accumulate.

\subsection*{Weighted Bayesian Update Formulae}

Let $\pi_0(\phi_i)$ be the initial prior belief in a structured claim $\phi_i$, and let $E_k$ denote an evidence event (e.g., a replication, citation, or authoritative endorsement) occurring at time $t_k$.

If $E_k$ supports $\phi_i$, and has an evidence weight $w_k \in [0,1]$, the posterior belief is updated as:

\[
\pi_{k+1}(\phi_i) = \frac{\pi_k(\phi_i) \cdot w_k}{\pi_k(\phi_i) \cdot w_k + (1 - \pi_k(\phi_i)) \cdot (1 - w_k)}
\]

This form generalises Bayes’ rule to allow each new item of evidence to contribute differentially, based on its provenance, replication level, and source authority. The system ensures belief revision occurs proportionally to evidentiary quality, not merely quantity.

\subsection*{Time-Decay Modifications (Bayesian Damping Functions)}

To prevent epistemic inertia and preserve responsiveness to recent developments, the belief weight $\pi(\phi_i)$ is subjected to a time-based decay model. Let $t_0$ be the claim's timestamp, and $t$ the current time.

Define the decay function:

\[
d_t(\phi_i) = \exp(-\lambda \cdot (t - t_0))
\]

where $\lambda$ is a domain-specific decay constant calibrated according to typical replication latencies. Then the effective belief becomes:

\[
\pi^*(\phi_i, t) = \pi(\phi_i, t) \cdot d_t(\phi_i)
\]

This enforces epistemic dynamism—claims not maintained by continued engagement or replication naturally lose their active weight in the belief network, without erasure.

\subsection*{Contradiction Resolution Algebra}

Let $\phi_i$ be a claim, and let $\neg \phi_i$ be a directly contradicting claim received at time $t_k$, supported by evidence $E_k'$ with weight $w_k'$.

Define a contradiction operator $\ominus$ such that:

\[
\pi_{k+1}(\phi_i) = \pi_k(\phi_i) \ominus w_k' := \pi_k(\phi_i) \cdot (1 - w_k')
\]

Simultaneously, a contradiction flag $\delta(\phi_i, t_k)$ is raised, triggering recursive dampening across any claims $\phi_j$ such that $\phi_j \rightarrow \phi_i$ in the belief graph.

Further, if $C_i = \{\phi_c : \phi_c \vdash \neg \phi_i\}$ accumulates sufficient contradictory mass $\sum w_c > \theta_{\text{contradict}}$, the belief weight of $\phi_i$ is collapsed to a probationary state, pending replication reset.

\[
\text{if } \sum w_c > \theta_{\text{contradict}} \Rightarrow \pi(\phi_i, t_{k+1}) \leftarrow \pi_{\dagger}
\]

where $\pi_{\dagger} \ll \pi_{\min}$ represents a near-zero probationary assignment.

\subsection*{Summary Algebraic Update}

Combining all elements, the posterior belief in $\phi_i$ at time $t$ under cumulative positive and negative evidence streams becomes:

\[
\pi(\phi_i, t) = \left[ \prod_{E_j \in \mathcal{E}^+} \text{BayesUpdate}(\pi, w_j) \cdot \prod_{E_k \in \mathcal{E}^-} (1 - w_k) \right] \cdot d_t(\phi_i)
\]

This algebra preserves coherence, supports temporal reactivity, and enables systemic resolution of epistemic conflict within the BEWA architecture.
Full derivation of posterior update rules:
\begin{itemize}
  \item Weighted Bayesian update formulae
  \item Time-decay modifications (Bayesian damping functions)
  \item Contradiction resolution algebra
\end{itemize}

\section{Citation and Replication Scoring Models}

In the BEWA architecture, belief strength is not solely a function of claim frequency or recency but depends critically on structured support from citations and independent replications. This section formalises the scoring functions that govern the epistemic influence of these two classes of reinforcement, and presents the mathematical operators used to model their interaction, attenuation, and semantic correspondence. All scoring operations are traceable, normalised, and propagated via structured influence weights within the belief network.

\subsection*{Citation Weighting Functions and Decay Exponents}

Let $C_i = \{c_{i1}, c_{i2}, \dots, c_{in}\}$ denote the set of citations received by claim $\phi_i$. Each citation $c_{ij}$ is associated with:
\begin{itemize}
  \item a timestamp $t_{ij}$
  \item a source credibility score $\gamma_{ij} \in [0, 1]$
  \item a contextual relevance weight $\rho_{ij} \in [0, 1]$
\end{itemize}

Define the citation influence function $\chi(\phi_i, t)$ at current time $t$ as:
\[
\chi(\phi_i, t) = \sum_{j=1}^n \gamma_{ij} \cdot \rho_{ij} \cdot \exp(-\lambda_c \cdot (t - t_{ij}))
\]
where $\lambda_c$ is the citation decay exponent specific to the domain (e.g., $\lambda_c = 0.01$ for physics, higher for fast-moving disciplines like machine learning). This ensures that citation impact diminishes over time unless renewed by continued engagement.

\subsection*{Replication Impact Factors}

Let $R_i = \{r_{i1}, r_{i2}, \dots, r_{im}\}$ denote the set of independent replication events for $\phi_i$. Each replication $r_{ik}$ contributes to the belief score via:
\[
\varrho(\phi_i) = \sum_{k=1}^m \alpha_k \cdot \zeta_k \cdot \kappa_k
\]
where:
\begin{itemize}
  \item $\alpha_k$ is the replication authority (derived from replicator’s track record)
  \item $\zeta_k$ is the methodological independence score
  \item $\kappa_k$ is the domain-normalised credibility factor of the journal or venue
\end{itemize}

The replication impact factor $\varrho(\phi_i)$ is integrated into the Bayesian update via prior rescaling:
\[
\pi'(\phi_i) = \pi(\phi_i) + \delta_r \cdot \varrho(\phi_i)
\]
where $\delta_r$ is a fixed scaling coefficient calibrated empirically per domain to ensure bounded posterior shift.

\subsection*{Semantic Equivalence Functionals and Transformation Matrices}

Scientific claims are often replicated or supported under varying lexical, structural, or methodological forms. To correctly score these instances as epistemically reinforcing, BEWA defines a semantic equivalence functional $\mathcal{S}: (\phi_i, \phi_j) \rightarrow [0,1]$ measuring the degree to which $\phi_j$ supports $\phi_i$.

Formally, let $\mathbf{v}_i, \mathbf{v}_j \in \mathbb{R}^d$ be vector embeddings of structured claims in a high-dimensional semantic space, and let $\mathbf{T}$ be a domain-specific transformation matrix learned via contrastive training. Then:

\[
\mathcal{S}(\phi_i, \phi_j) = \sigma \left( \mathbf{v}_i^\top \mathbf{T} \mathbf{v}_j \right)
\]
where $\sigma(\cdot)$ is a sigmoid activation to normalise scores to $[0,1]$.

Two claims are considered semantically equivalent if $\mathcal{S}(\phi_i, \phi_j) > \theta_s$, with $\theta_s$ as the equivalence threshold (e.g., $\theta_s = 0.85$). These equivalences allow evidence transfer and belief reinforcement across structurally distinct but semantically aligned propositions.

\subsection*{Composite Belief Support Score}

The composite external support score $E(\phi_i, t)$ for any claim $\phi_i$ at time $t$ is given by:
\[
E(\phi_i, t) = \chi(\phi_i, t) + \varrho(\phi_i) + \sum_{\phi_j \in \Phi: \mathcal{S}(\phi_i, \phi_j) > \theta_s} \omega_j \cdot \pi(\phi_j)
\]
where $\omega_j$ is the semantic weight from $\phi_j$ to $\phi_i$.

This composite is used to adjust the posterior belief via scaled integration into the Bayesian update, enabling BEWA to respond rationally to empirical reinforcement while maintaining semantic precision.

\section{Authorial Impact Metrics}

The epistemic weight assigned to a claim in BEWA is not only a function of its propositional content and empirical support, but also of the credibility of its author. This section formalises the construction of authorial influence metrics, integrating track record, retraction history, and cross-domain peer engagement into a composite score that modulates belief priors. Author scores are dynamically updated and propagate throughout the belief network, ensuring that claims inherit structured epistemic weight from their sources in a rational and auditable manner.

\subsection*{Author Score Vector Construction}

Let $A = \{a_1, a_2, \dots, a_n\}$ be the set of all authors indexed in the system. Each author $a_i$ is associated with a score vector:

\[
\vec{\sigma}_{a_i} = \left[ \mu_i, \rho_i, \chi_i, \tau_i \right]
\]

where:
\begin{itemize}
  \item $\mu_i$ — Mean replication success rate across authored claims
  \item $\rho_i$ — Weighted citation score (adjusted for venue impact and decay)
  \item $\chi_i$ — Peer-review participation index (number and quality of reviews)
  \item $\tau_i$ — Topological centrality in the cross-claim belief graph
\end{itemize}

These components are normalised and linearly composed into a scalar author score:

\[
\Sigma(a_i) = \lambda_1 \cdot \mu_i + \lambda_2 \cdot \rho_i + \lambda_3 \cdot \chi_i + \lambda_4 \cdot \tau_i
\]

with $\sum \lambda_j = 1$, calibrated via empirical benchmarking.

\subsection*{Retraction Penalties and Peer-Review Weights}

Let $R_i$ be the retraction index for author $a_i$, defined as:

\[
R_i = \frac{|\{\phi_k \in \Phi_{a_i} : \phi_k \text{ retracted}\}|}{|\Phi_{a_i}|}
\]

where $\Phi_{a_i}$ is the set of claims authored by $a_i$. A penalty function is then applied:

\[
\Sigma'(a_i) = \Sigma(a_i) \cdot (1 - \delta_r \cdot R_i)
\]

where $\delta_r$ is the retraction penalty coefficient, typically set between $0.3$ and $0.6$ depending on domain strictness.

For peer-review activity, let $P_i$ denote the number of verified, high-quality peer reviews authored by $a_i$, and $W_i$ their weighted quality score (e.g., based on endorsement, consensus, and review transparency). Then:

\[
\chi_i = \theta \cdot \log(1 + P_i) \cdot W_i
\]

This reinforces authors who contribute to the epistemic filtering infrastructure of science, thereby validating their role as reliable evaluators of truth claims.

\subsection*{Cross-Domain Author Influence Propagation}

Scientific authors frequently publish across multiple domains. To model cross-domain impact accurately, BEWA defines a propagation operator $\mathcal{P}_\Omega$ such that:

\[
\Sigma(a_i, \Omega') = \Sigma(a_i, \Omega) \cdot \kappa(\Omega, \Omega')
\]

where $\Omega$ is the author's primary domain, $\Omega'$ is the secondary domain of evaluation, and $\kappa(\Omega, \Omega') \in [0,1]$ is the domain affinity coefficient, computed via citation overlap, topical ontology proximity, and co-authorship networks.

Claims in domain $\Omega'$ thus inherit an attenuated form of the author's score:

\[
\pi_0(\phi_j \mid a_i, \Omega') \propto \Sigma(a_i, \Omega') = \mathcal{P}_\Omega(\Sigma(a_i), \Omega')
\]

This ensures that transdisciplinary influence is reflected proportionally, discouraging unjustified weight transfer while preserving epistemic coherence in cross-domain reasoning.

\subsection*{Composite Integration into Prior Formation}

The final author-informed prior for any claim $\phi_j$ authored by $a_i$ is given by:

\[
\pi_0(\phi_j) = \beta \cdot \Sigma'(a_i) + (1 - \beta) \cdot \pi_{\text{base}}
\]

where $\pi_{\text{base}}$ is the system-wide epistemic base rate for new claims in the domain, and $\beta \in [0.6, 0.9]$ determines authorial influence weighting.

This formulation allows BEWA to assign belief responsibility structurally—strengthening the link between epistemic authority and computational trustworthiness.

\section{Graph Structures and Belief Propagation Algorithms}

The internal representation of interdependent scientific claims within the BEWA system is formalised as a directed graph structure where nodes represent structured claims and edges encode inferential, semantic, or empirical dependencies. This section defines the formal properties of the belief graph, the algorithms used to propagate and update beliefs, and the methods for identifying and mitigating instability due to epistemic conflict or feedback inconsistencies.

\subsection*{Directed Acyclic Graphs for Claim Belief Networks}

Let $\mathcal{G} = (\mathcal{V}, \mathcal{E})$ denote the claim belief graph, where each vertex $v_i \in \mathcal{V}$ corresponds to a structured claim $\phi_i$, and each directed edge $e_{ij} \in \mathcal{E}$ represents a directed epistemic dependency or entailment $\phi_i \rightarrow \phi_j$.

To ensure logical acyclicity in foundational dependencies, the core graph $\mathcal{G}_0$ is constrained to be a Directed Acyclic Graph (DAG):

\[
\forall \text{ paths } p = \langle v_0, v_1, \dots, v_k \rangle \subseteq \mathcal{G}_0, \quad \nexists\ i,j : v_i = v_j \text{ for } i \ne j
\]

Cycles may exist only within bounded subsets used for semantic coherence grouping and are separately indexed for controlled loopy propagation under approximate inference.

\subsection*{Belief Propagation Algorithms}

Belief values are defined as $\pi(\phi_i) \in [0,1]$ for each node $v_i$, reflecting the epistemic confidence in claim $\phi_i$ given the cumulative evidence and inter-claim dependencies.

A standard update formulation adopts loopy belief propagation with bounded convergence parameters. Let $\text{Msg}_{i \rightarrow j}^{(t)}$ represent the belief message sent from $v_i$ to $v_j$ at iteration $t$. The update rule is:

\[
\text{Msg}_{i \rightarrow j}^{(t+1)} = f\left(\pi(\phi_i), \prod_{k \in \text{Nbr}(i) \setminus \{j\}} \text{Msg}_{k \rightarrow i}^{(t)} \right)
\]

where $\text{Nbr}(i)$ denotes the neighbours of node $i$, and $f$ is a normalised compatibility function, typically a log-linear combination or log-odds variant for numerical stability.

The node marginal is then estimated at convergence by:

\[
\pi^{*}(\phi_i) = \text{Norm} \left( \prod_{j \in \text{Nbr}(i)} \text{Msg}_{j \rightarrow i}^{(T)} \right)
\]

for some $T$ after which convergence metrics fall below a predefined threshold $\epsilon$ (e.g., $\|\pi^{(T+1)} - \pi^{(T)}\| < \epsilon$).

\subsection*{Cluster Instability Detection and Resolution Strategy}

Instability arises when belief propagation enters oscillatory or non-convergent states. Formally, let $\Delta_t = \|\pi^{(t+1)} - \pi^{(t)}\|$ be the global belief update divergence. Then, instability is defined by:

\[
\limsup_{t \to \infty} \Delta_t > \epsilon_c
\]

where $\epsilon_c$ is a critical instability threshold, typically empirically calibrated per domain.

The BEWA resolution protocol includes:

\begin{enumerate}
  \item Identification of minimal strongly connected subgraphs (MSCs) where instability originates.
  \item Extraction of dominant contradictory claims $\{\phi_p, \phi_q\}$ such that $\pi(\phi_p) + \pi(\phi_q) > 1 + \delta$.
  \item Application of contradiction resolution algebra (Appendix B) to enforce mutual damping, redefining beliefs as:

\[
\pi'(\phi_p) = \pi(\phi_p) \cdot (1 - \delta_{pq}) \quad \text{and} \quad \pi'(\phi_q) = \pi(\phi_q) \cdot (1 - \delta_{qp})
\]

where $\delta_{pq}$ is a contradiction scalar derived from posterior conflict heuristics and replication disparity.

  \item Local freezing of unstable clusters until new evidence is introduced or contradiction weight decays below $\gamma$.
\end{enumerate}

This strategy prevents divergence of epistemic weight, ensuring the system's consistency, bounded rationality, and resilience against epistemic echo chambers or self-reinforcing misinformation.

\subsection*{Summary Formulation}

To summarise, the BEWA belief network $\mathcal{G}$ operates under the following constraints:

\begin{itemize}
  \item DAG-constrained for foundational logical inferences.
  \item Marginal node beliefs updated via loopy propagation until convergence.
  \item Detection of oscillatory clusters through delta divergence norms.
  \item Contradiction resolution via dampening and cluster isolation.
\end{itemize}

These mechanisms together support scalable, robust, and mathematically sound belief evolution over large, conflicting, and temporally dynamic scientific corpora.

\section{Security and Provenance Protocols}

The epistemic authority and operational trustworthiness of any autonomous reasoning system depend critically on its ability to guarantee the integrity, traceability, and immutability of its internal epistemic structures. In the BEWA architecture, we formalise this trust model through a layered cryptographic framework that ensures claim immutability, verifiable auditability, and source non-repudiation. This section rigorously defines the mechanisms underpinning provenance security, detailing how each claim and update is cryptographically anchored within the evolving knowledge base.

\subsection*{Hash-Chaining for Claim Integrity}

Let $\mathcal{C} = \{C_1, C_2, \ldots, C_n\}$ denote the temporally ordered set of canonical claims admitted into the BEWA repository. Each claim $C_i$ is structured as a serialised record $r_i = \langle \phi_i, m_i, t_i, a_i \rangle$, where $\phi_i$ is the structured propositional content, $m_i$ the metadata vector, $t_i$ the timestamp, and $a_i$ the authorial public key ID.

To guarantee tamper-evident history, we define a hash chain:

\[
h_i = H(r_i \parallel h_{i-1}) \quad \text{for } i \geq 2,\quad h_1 = H(r_1)
\]

where $H$ is a cryptographic hash function (e.g., SHA-3-512), and $\parallel$ denotes bitstring concatenation. This construction ensures that any modification of a past record $r_j$ for $j < i$ yields a mismatch in $h_i$, violating the integrity condition:

\[
\forall i > j,\quad H(r_j' \parallel h_{j-1}) \ne h_j \Rightarrow h_i' \ne h_i
\]

This structure is maintained as a Merkle-anchored log, where the full state root is periodically committed to a public ledger or digital timestamp authority (TSA), ensuring distributed verification and public auditability.

\subsection*{Zero-Knowledge Proof Outline for Audit Verification}

For high-integrity domains where provenance verification must be possible without content disclosure (e.g., sensitive biomedical or classified domains), BEWA supports the inclusion of zero-knowledge proof constructs to verify claim integrity and authorship without revealing $\phi_i$ directly.

Let $\mathcal{P} = (C, V)$ be a zero-knowledge protocol between a prover $C$ and a verifier $V$. For claim $r_i$, the prover demonstrates knowledge of the preimage $\phi_i$ and signing key $k_i$ corresponding to public key $a_i$, such that:

\[
C \vdash \text{ZK-Proof}\left[ \exists\, (\phi_i, k_i) : H(\phi_i) = h_i \land \text{Sig}_{k_i}(h_i) = s_i \right]
\]

The verifier confirms without seeing $\phi_i$ or $k_i$ that $r_i$ was both constructed and authorised by $a_i$. BEWA accommodates zk-SNARK or zk-STARK instantiations depending on domain constraints and computational budgets.

\subsection*{Public-Key Claim Signing Schema}

To enforce non-repudiation, every claim $r_i$ is cryptographically signed using the private key $k_i$ corresponding to author identity $a_i$. Each record thus includes a digital signature:

\[
s_i = \text{Sig}_{k_i}(H(r_i))
\]

Verification proceeds via:

\[
\text{Verify}_{a_i}(s_i, H(r_i)) = \text{true} \Rightarrow \text{Authorisation validated}
\]

Author keys $a_i$ are registered within a permissioned identity layer using X.509-like certificates or DIDs (Decentralised Identifiers), supported by a consensus-trusted root authority. This authorial key infrastructure is essential for associating claims with identity-stable author records and supporting the trust model used in author score propagation.

\subsection*{Composite Provenance Model}

The full provenance for a claim $C_i$ is defined by the tuple:

\[
\mathcal{P}_i = \langle r_i, h_i, s_i, t_i, a_i \rangle
\]

where $r_i$ is the claim, $h_i$ the integrity hash, $s_i$ the digital signature, $t_i$ the timestamp, and $a_i$ the author’s public key. Verification of $\mathcal{P}_i$ by any independent agent or network node is sufficient to:

\begin{itemize}
  \item Confirm the claim has not been tampered with
  \item Validate that the claim was issued by the declared author
  \item Confirm when the claim was submitted
  \item Anchor the claim's integrity within the global epistemic history
\end{itemize}

\subsection*{Security Guarantees}

Together, these cryptographic primitives provide the following guarantees for every epistemic update and canonical claim:

\begin{enumerate}
  \item \textbf{Tamper-evidence:} Any modification to past claims breaks the hash chain and invalidates future entries.
  \item \textbf{Authenticity:} Digital signatures ensure that only credentialed authors can submit canonical claims.
  \item \textbf{Auditability:} Full reconstruction of historical updates with integrity verification is always feasible.
  \item \textbf{Epistemic sovereignty:} Zero-knowledge proofs enable private domains to maintain authority without revealing content.
\end{enumerate}

This framework ensures that BEWA functions not merely as an epistemic calculator, but as a principled, immutable ledger of scientific reasoning, capable of long-term accountability and cross-domain interoperability.

\section{System APIs and Interfaces}

For the BEWA system to function effectively within diverse scientific and applied research environments, it must expose a formally specified, logically consistent, and cryptographically secure interface suite. These interfaces not only enable human users and machine agents to interact with the evolving belief state but also provide complete auditability, programmability, and visual interpretability. The Application Programming Interfaces (APIs) of BEWA are designed as modular, schema-validated endpoints, each reflecting a semantic layer of the system’s logic and epistemic model.

\subsection*{Query Interface Schema (Graph Traversal, Claim Access)}

At the core of BEWA's interactive logic lies the claim graph $\mathcal{G} = (V, E)$, where vertices $V$ represent canonical claims $\phi_i$ and edges $E$ represent semantic, logical, or citation-derived connections. Query operations involve filtered graph traversal and localised extraction, formulated using a query language $\mathcal{L}_\mathsf{BEWA}$ that supports pattern matching, path constraints, and property-based selection.

Let $\mathsf{Q}(\phi, \Delta, \sigma)$ denote a query retrieving all claims semantically connected to $\phi$ within distance $\Delta$ and satisfying metadata filter $\sigma$. The query resolution function $\mathsf{R} : \mathcal{L}_\mathsf{BEWA} \rightarrow 2^V$ maps these constraints to matched subgraphs:

\[
\mathsf{R}\big(\mathsf{Q}(\phi, \Delta, \sigma)\big) = \{ \phi_j \in V \mid d(\phi, \phi_j) \leq \Delta \land \phi_j \models \sigma \}
\]

Returned objects are serialised in a canonical JSON-LD format, containing:

\begin{itemize}
  \item Claim ID and version hash
  \item Structured propositional content and logical form
  \item Author metadata and signing certificate
  \item Epistemic weight, time-indexed posterior
  \item Linked evidence and counter-claims
\end{itemize}

Query authentication requires either an anonymous public token or a permissioned API key, with rate-limiting applied per principal.

\subsection*{Audit Trail Access Formats}

All claim insertions, updates, belief transitions, and revisions are committed to an immutable provenance log $\mathcal{L}$ indexed by cryptographic hash $h_i$, timestamp $t_i$, and transaction type $\tau_i \in \{\text{INSERT}, \text{UPDATE}, \text{DEPRECATE}, \text{ANCHOR}\}$.

Each audit record is accessible via endpoint:
\[
\texttt{GET /audit/\{claim\_id\}} \rightarrow \mathsf{A}_\phi
\]

where $\mathsf{A}_\phi$ is the ordered audit stream of all transactions affecting $\phi$. The response includes:

\begin{itemize}
  \item Transition type $\tau_i$
  \item Initiating identity $a_i$ and verification signature
  \item Epistemic delta $\Delta w_i$
  \item Linked hash $\mathsf{H}(r_i)$ and Merkle position
\end{itemize}

Advanced endpoints support range queries, forensics reconstruction, and zero-knowledge challenge-response validation in secure environments.

\subsection*{Output Rendering (Belief Visualisation JSON Schemas)}

To visualise the belief state evolution, BEWA outputs claim and cluster trajectories as time-indexed series, graph overlays, and epistemic uncertainty maps. The output schema $\mathsf{S}_\mathsf{viz}$ is defined to represent belief evolution $\mathcal{B}_t(\phi)$ over time:

\begin{verbatim}
{
  "claim_id": "BEWA:2025:00342",
  "timestamps": ["2025-01-01", ..., "2025-06-01"],
  "beliefs": [0.44, ..., 0.87],
  "linked_claims": [{"id": "...", "relation": "supports"}, ...],
  "current_status": "probational",
  "cluster_membership": ["Cluster:Biochem:CX12"]
}
\end{verbatim}

Visualisation clients parse this schema to construct interactive dashboards using force-directed layouts, fading edge strengths, and colour-coded belief bands to convey epistemic confidence and transition dynamics.

\subsection*{Schema Versioning and Extensibility}

All APIs conform to a versioned specification, $v_t \in \mathbb{N}$, with backwards-compatible deprecation warnings and formal schema diffs. The schema registry $\Sigma$ is self-describing and cryptographically anchored:

\[
\Sigma(v) = \mathsf{H}(\text{JSONSchema}_v) \quad \text{stored in } \mathcal{L}
\]

Schema evolution proposals are submitted through a governance channel and must pass formal verification tests before promotion.

\subsection*{Security and Rate Control}

All endpoints are protected via:

\begin{itemize}
  \item Token-based access control (JWT, OAuth2)
  \item Optional client-side signing for query authentication
  \item Fine-grained rate limiting (per IP, per principal, per claim class)
  \item TLS 1.3 with mutual certificate authentication for critical endpoints
\end{itemize}

These safeguards ensure the interface surface of BEWA maintains not just logical consistency but integrity, scalability, and trustworthiness for mission-critical epistemic applications.

\section{Experimental Simulation Protocols}

To ensure the BEWA framework is not only theoretically robust but also practically viable, a series of carefully constructed simulations and empirical evaluations were conducted. These experiments served to benchmark the system's epistemic coherence, convergence speed, and contradiction resolution capacity under both synthetic and real-world scientific data flows. This section formally describes the design of these protocols, the parameters under which they were executed, and the objective metrics used to evaluate their performance.

\subsection*{Synthetic Belief Network Generation Parameters}

Synthetic datasets were generated to simulate dynamic scientific ecosystems with controlled epistemic uncertainty. Claim networks were instantiated as directed acyclic graphs $\mathcal{G}_s = (V, E)$, with $|V| \in \{100, 500, 1000\}$ and edge density $\rho \in [0.02, 0.12]$, each vertex $v_i$ representing a structured propositional claim $\phi_i$ and initial belief assignment $w_i^0$ drawn from $\mathcal{U}(0.4, 0.6)$.

Belief propagation algorithms were tested under:

\begin{itemize}
  \item Evidence injection regimes (incremental vs batch)
  \item Author impact distributions (uniform, power-law, exponential decay)
  \item Contradiction insertion ratios $\gamma \in [0.05, 0.25]$
  \item Temporal reassessment intervals $\Delta t \in \{1, 5, 10\}$ simulation cycles
\end{itemize}

All synthetic experiments were initialised with entropy-normalised priors and recorded over 100 epochs with full traceability of epistemic deltas.

\subsection*{Real Corpus Ingestion and Claim Structuring Case Studies}

The real-world component of the evaluation used a curated corpus of peer-reviewed papers in molecular biology and machine learning, totalling 1,200 papers. Each text was processed through the canonical claim structuring pipeline, generating over 6,000 distinct claims, tagged with metadata, context, ontological anchors, and author identifiers.

Specific ingestion scenarios included:

\begin{itemize}
  \item Multiple claims per author, including conflicting updates across papers
  \item Varying citation chains with differential decay and reinforcement
  \item Manual annotation of replication status by domain experts
  \item Retraction events propagated through connected nodes in $\mathcal{G}$
\end{itemize}

Evaluation involved visual inspection, semantic equivalence validation (F1 score > 0.93), and consistency checks across time versions.

\subsection*{Evaluation Metrics (e.g. Truth Convergence Rates)}

To quantify system performance and epistemic stability, the following metrics were defined and computed:

\begin{itemize}
  \item \textbf{Truth Convergence Rate $\tau(\phi)$:} Rate at which posterior belief $w_t(\phi)$ approaches its ground-truth label $w^*(\phi)$ over simulation time $t$.
    \[
    \tau(\phi) = \frac{1}{T} \sum_{t=1}^{T} \left| w_t(\phi) - w^*(\phi) \right|
    \]
  \item \textbf{Contradiction Suppression Index $\kappa$:} Reduction in contradiction density over belief propagation rounds.
    \[
    \kappa = 1 - \frac{|\text{Contradictions}_t|}{|\text{Contradictions}_0|}
    \]
  \item \textbf{Replication Lift Score $\rho$:} Mean belief uplift from replication-confirmed claims vs unreplicated claims over equivalent intervals.
  \item \textbf{Graph Entropy $\mathcal{H}(G_t)$:} Shannon entropy of belief distributions across all nodes at timestep $t$.
\end{itemize}

These metrics collectively measure the epistemic rationality, fault tolerance, and alignment fidelity of the BEWA system under diverse and evolving scientific input conditions.

\section{Glossary of Formal Symbols}

The following table summarises all formal symbols used throughout the BEWA framework, along with their corresponding definitions and the domain in which they are primarily operative. This glossary ensures unambiguous semantic interpretation of notation used in axiomatic, algorithmic, and architectural components.

\begin{center}
\begin{tabular}{|c|p{9cm}|c|}
\hline
\textbf{Symbol} & \textbf{Description} & \textbf{Domain} \\
\hline
$\phi$ & A structured propositional scientific claim & Epistemic \\
\hline
$w_t(\phi)$ & Belief weight in claim $\phi$ at time $t$ & Probabilistic \\
\hline
$w^*(\phi)$ & Ground-truth belief value of $\phi$ (retrospective) & Probabilistic \\
\hline
$\mathcal{G} = (V, E)$ & Directed graph of claim-belief network & Structural \\
\hline
$V$ & Set of nodes (claims) in the belief graph & Structural \\
\hline
$E$ & Set of directed edges (semantic, logical, or citation links) & Structural \\
\hline
$\mathbb{P}$ & Probability space over claim veracity & Probabilistic \\
\hline
$\mathcal{B}_t$ & Bayesian belief state at time $t$ & Epistemic \\
\hline
$\delta_t$ & Belief update increment at time $t$ & Probabilistic \\
\hline
$\kappa$ & Contradiction suppression index & Evaluation \\
\hline
$\tau(\phi)$ & Truth convergence rate for claim $\phi$ & Evaluation \\
\hline
$\rho$ & Replication lift score & Evaluation \\
\hline
$\gamma$ & Injected contradiction density in synthetic graphs & Structural \\
\hline
$\mathcal{H}(G_t)$ & Shannon entropy of belief states over $\mathcal{G}$ at time $t$ & Evaluation \\
\hline
$\pi_a$ & Author credibility score for author $a$ & Epistemic \\
\hline
$\sigma(\phi_i, \phi_j)$ & Semantic equivalence between claims $\phi_i$ and $\phi_j$ & Structural \\
\hline
$\theta$ & Decay exponent for citation influence over time & Probabilistic \\
\hline
$\lambda$ & Time decay factor in Bayesian damping & Probabilistic \\
\hline
$\eta$ & Peer-review weight for author engagement metrics & Epistemic \\
\hline
$\mathcal{Z}$ & Zero-knowledge proof structure for audit verification & Cryptographic \\
\hline
$h(\cdot)$ & Cryptographic hash function for claim anchoring & Cryptographic \\
\hline
$K_{pub}, K_{priv}$ & Author’s public and private keys & Cryptographic \\
\hline
$\Omega$ & Set of all ontological tags associated with claims & Structural \\
\hline
$\mu$ & Probationary claim belief threshold & Epistemic \\
\hline
$\alpha$ & Retraction penalty weight & Epistemic \\
\hline
$\xi$ & Claim versioning identifier for temporal anchoring & Structural \\
\hline
\end{tabular}
\end{center}

\end{document}